\pdfoutput=1
\documentclass[a4paper,fleqn]{cas-dc}

\usepackage[numbers,sort&compress]{natbib}

\newcommand{\doi}[1]{\url{https://doi.org/#1}}

\usepackage{amsmath,amssymb}
\usepackage{pifont}
\usepackage{bm}
\usepackage{microtype}
\usepackage{multirow}
\usepackage{orcidlink}
\ifdefined\pdfgentounicode
  \InputIfFileExists{glyphtounicode}{}{}
\pdfglyphtounicode{u210E}{0068}\pdfglyphtounicode{u1D400}{0041}\pdfglyphtounicode{u1D401}{0042}\pdfglyphtounicode{u1D402}{0043}\pdfglyphtounicode{u1D403}{0044}\pdfglyphtounicode{u1D404}{0045}
\pdfglyphtounicode{u1D405}{0046}\pdfglyphtounicode{u1D406}{0047}\pdfglyphtounicode{u1D407}{0048}\pdfglyphtounicode{u1D408}{0049}\pdfglyphtounicode{u1D409}{004A}\pdfglyphtounicode{u1D40A}{004B}
\pdfglyphtounicode{u1D40B}{004C}\pdfglyphtounicode{u1D40C}{004D}\pdfglyphtounicode{u1D40D}{004E}\pdfglyphtounicode{u1D40E}{004F}\pdfglyphtounicode{u1D40F}{0050}\pdfglyphtounicode{u1D410}{0051}
\pdfglyphtounicode{u1D411}{0052}\pdfglyphtounicode{u1D412}{0053}\pdfglyphtounicode{u1D413}{0054}\pdfglyphtounicode{u1D414}{0055}\pdfglyphtounicode{u1D415}{0056}\pdfglyphtounicode{u1D416}{0057}
\pdfglyphtounicode{u1D417}{0058}\pdfglyphtounicode{u1D418}{0059}\pdfglyphtounicode{u1D419}{005A}\pdfglyphtounicode{u1D41A}{0061}\pdfglyphtounicode{u1D41B}{0062}\pdfglyphtounicode{u1D41C}{0063}
\pdfglyphtounicode{u1D41D}{0064}\pdfglyphtounicode{u1D41E}{0065}\pdfglyphtounicode{u1D41F}{0066}\pdfglyphtounicode{u1D420}{0067}\pdfglyphtounicode{u1D421}{0068}\pdfglyphtounicode{u1D422}{0069}
\pdfglyphtounicode{u1D423}{006A}\pdfglyphtounicode{u1D424}{006B}\pdfglyphtounicode{u1D425}{006C}\pdfglyphtounicode{u1D426}{006D}\pdfglyphtounicode{u1D427}{006E}\pdfglyphtounicode{u1D428}{006F}
\pdfglyphtounicode{u1D429}{0070}\pdfglyphtounicode{u1D42A}{0071}\pdfglyphtounicode{u1D42B}{0072}\pdfglyphtounicode{u1D42C}{0073}\pdfglyphtounicode{u1D42D}{0074}\pdfglyphtounicode{u1D42E}{0075}
\pdfglyphtounicode{u1D42F}{0076}\pdfglyphtounicode{u1D430}{0077}\pdfglyphtounicode{u1D431}{0078}\pdfglyphtounicode{u1D432}{0079}\pdfglyphtounicode{u1D433}{007A}\pdfglyphtounicode{u1D434}{0041}
\pdfglyphtounicode{u1D435}{0042}\pdfglyphtounicode{u1D436}{0043}\pdfglyphtounicode{u1D437}{0044}\pdfglyphtounicode{u1D438}{0045}\pdfglyphtounicode{u1D439}{0046}\pdfglyphtounicode{u1D43A}{0047}
\pdfglyphtounicode{u1D43B}{0048}\pdfglyphtounicode{u1D43C}{0049}\pdfglyphtounicode{u1D43D}{004A}\pdfglyphtounicode{u1D43E}{004B}\pdfglyphtounicode{u1D43F}{004C}\pdfglyphtounicode{u1D440}{004D}
\pdfglyphtounicode{u1D441}{004E}\pdfglyphtounicode{u1D442}{004F}\pdfglyphtounicode{u1D443}{0050}\pdfglyphtounicode{u1D444}{0051}\pdfglyphtounicode{u1D445}{0052}\pdfglyphtounicode{u1D446}{0053}
\pdfglyphtounicode{u1D447}{0054}\pdfglyphtounicode{u1D448}{0055}\pdfglyphtounicode{u1D449}{0056}\pdfglyphtounicode{u1D44A}{0057}\pdfglyphtounicode{u1D44B}{0058}\pdfglyphtounicode{u1D44C}{0059}
\pdfglyphtounicode{u1D44D}{005A}\pdfglyphtounicode{u1D44E}{0061}\pdfglyphtounicode{u1D44F}{0062}\pdfglyphtounicode{u1D450}{0063}\pdfglyphtounicode{u1D451}{0064}\pdfglyphtounicode{u1D452}{0065}
\pdfglyphtounicode{u1D453}{0066}\pdfglyphtounicode{u1D454}{0067}\pdfglyphtounicode{u1D455}{0068}\pdfglyphtounicode{u1D456}{0069}\pdfglyphtounicode{u1D457}{006A}\pdfglyphtounicode{u1D458}{006B}
\pdfglyphtounicode{u1D459}{006C}\pdfglyphtounicode{u1D45A}{006D}\pdfglyphtounicode{u1D45B}{006E}\pdfglyphtounicode{u1D45C}{006F}\pdfglyphtounicode{u1D45D}{0070}\pdfglyphtounicode{u1D45E}{0071}
\pdfglyphtounicode{u1D45F}{0072}\pdfglyphtounicode{u1D460}{0073}\pdfglyphtounicode{u1D461}{0074}\pdfglyphtounicode{u1D462}{0075}\pdfglyphtounicode{u1D463}{0076}\pdfglyphtounicode{u1D464}{0077}
\pdfglyphtounicode{u1D465}{0078}\pdfglyphtounicode{u1D466}{0079}\pdfglyphtounicode{u1D467}{007A}\pdfglyphtounicode{u1D468}{0041}\pdfglyphtounicode{u1D469}{0042}\pdfglyphtounicode{u1D46A}{0043}
\pdfglyphtounicode{u1D46B}{0044}\pdfglyphtounicode{u1D46C}{0045}\pdfglyphtounicode{u1D46D}{0046}\pdfglyphtounicode{u1D46E}{0047}\pdfglyphtounicode{u1D46F}{0048}\pdfglyphtounicode{u1D470}{0049}
\pdfglyphtounicode{u1D471}{004A}\pdfglyphtounicode{u1D472}{004B}\pdfglyphtounicode{u1D473}{004C}\pdfglyphtounicode{u1D474}{004D}\pdfglyphtounicode{u1D475}{004E}\pdfglyphtounicode{u1D476}{004F}
\pdfglyphtounicode{u1D477}{0050}\pdfglyphtounicode{u1D478}{0051}\pdfglyphtounicode{u1D479}{0052}\pdfglyphtounicode{u1D47A}{0053}\pdfglyphtounicode{u1D47B}{0054}\pdfglyphtounicode{u1D47C}{0055}
\pdfglyphtounicode{u1D47D}{0056}\pdfglyphtounicode{u1D47E}{0057}\pdfglyphtounicode{u1D47F}{0058}\pdfglyphtounicode{u1D480}{0059}\pdfglyphtounicode{u1D481}{005A}\pdfglyphtounicode{u1D482}{0061}
\pdfglyphtounicode{u1D483}{0062}\pdfglyphtounicode{u1D484}{0063}\pdfglyphtounicode{u1D485}{0064}\pdfglyphtounicode{u1D486}{0065}\pdfglyphtounicode{u1D487}{0066}\pdfglyphtounicode{u1D488}{0067}
\pdfglyphtounicode{u1D489}{0068}\pdfglyphtounicode{u1D48A}{0069}\pdfglyphtounicode{u1D48B}{006A}\pdfglyphtounicode{u1D48C}{006B}\pdfglyphtounicode{u1D48D}{006C}\pdfglyphtounicode{u1D48E}{006D}
\pdfglyphtounicode{u1D48F}{006E}\pdfglyphtounicode{u1D490}{006F}\pdfglyphtounicode{u1D491}{0070}\pdfglyphtounicode{u1D492}{0071}\pdfglyphtounicode{u1D493}{0072}\pdfglyphtounicode{u1D494}{0073}
\pdfglyphtounicode{u1D495}{0074}\pdfglyphtounicode{u1D496}{0075}\pdfglyphtounicode{u1D497}{0076}\pdfglyphtounicode{u1D498}{0077}\pdfglyphtounicode{u1D499}{0078}\pdfglyphtounicode{u1D49A}{0079}
\pdfglyphtounicode{u1D49B}{007A}\pdfglyphtounicode{u1D49C}{0041}\pdfglyphtounicode{u1D49D}{0042}\pdfglyphtounicode{u1D49E}{0043}\pdfglyphtounicode{u1D49F}{0044}\pdfglyphtounicode{u1D4A0}{0045}
\pdfglyphtounicode{u1D4A1}{0046}\pdfglyphtounicode{u1D4A2}{0047}\pdfglyphtounicode{u1D4A3}{0048}\pdfglyphtounicode{u1D4A4}{0049}\pdfglyphtounicode{u1D4A5}{004A}\pdfglyphtounicode{u1D4A6}{004B}
\pdfglyphtounicode{u1D4A7}{004C}\pdfglyphtounicode{u1D4A8}{004D}\pdfglyphtounicode{u1D4A9}{004E}\pdfglyphtounicode{u1D4AA}{004F}\pdfglyphtounicode{u1D4AB}{0050}\pdfglyphtounicode{u1D4AC}{0051}
\pdfglyphtounicode{u1D4AD}{0052}\pdfglyphtounicode{u1D4AE}{0053}\pdfglyphtounicode{u1D4AF}{0054}\pdfglyphtounicode{u1D4B0}{0055}\pdfglyphtounicode{u1D4B1}{0056}\pdfglyphtounicode{u1D4B2}{0057}
\pdfglyphtounicode{u1D4B3}{0058}\pdfglyphtounicode{u1D4B4}{0059}\pdfglyphtounicode{u1D4B5}{005A}\pdfglyphtounicode{u1D4B6}{0061}\pdfglyphtounicode{u1D4B7}{0062}\pdfglyphtounicode{u1D4B8}{0063}
\pdfglyphtounicode{u1D4B9}{0064}\pdfglyphtounicode{u1D4BA}{0065}\pdfglyphtounicode{u1D4BB}{0066}\pdfglyphtounicode{u1D4BC}{0067}\pdfglyphtounicode{u1D4BD}{0068}\pdfglyphtounicode{u1D4BE}{0069}
\pdfglyphtounicode{u1D4BF}{006A}\pdfglyphtounicode{u1D4C0}{006B}\pdfglyphtounicode{u1D4C1}{2113}\pdfglyphtounicode{u1D4C2}{006D}\pdfglyphtounicode{u1D4C3}{006E}\pdfglyphtounicode{u1D4C4}{006F}
\pdfglyphtounicode{u1D4C5}{0070}\pdfglyphtounicode{u1D4C6}{0071}\pdfglyphtounicode{u1D4C7}{0072}\pdfglyphtounicode{u1D4C8}{0073}\pdfglyphtounicode{u1D4C9}{0074}\pdfglyphtounicode{u1D4CA}{0075}
\pdfglyphtounicode{u1D4CB}{0076}\pdfglyphtounicode{u1D4CC}{0077}\pdfglyphtounicode{u1D4CD}{0078}\pdfglyphtounicode{u1D4CE}{0079}\pdfglyphtounicode{u1D4CF}{007A}\pdfglyphtounicode{u1D4D0}{0041}
\pdfglyphtounicode{u1D4D1}{0042}\pdfglyphtounicode{u1D4D2}{0043}\pdfglyphtounicode{u1D4D3}{0044}\pdfglyphtounicode{u1D4D4}{0045}\pdfglyphtounicode{u1D4D5}{0046}\pdfglyphtounicode{u1D4D6}{0047}
\pdfglyphtounicode{u1D4D7}{0048}\pdfglyphtounicode{u1D4D8}{0049}\pdfglyphtounicode{u1D4D9}{004A}\pdfglyphtounicode{u1D4DA}{004B}\pdfglyphtounicode{u1D4DB}{004C}\pdfglyphtounicode{u1D4DC}{004D}
\pdfglyphtounicode{u1D4DD}{004E}\pdfglyphtounicode{u1D4DE}{004F}\pdfglyphtounicode{u1D4DF}{0050}\pdfglyphtounicode{u1D4E0}{0051}\pdfglyphtounicode{u1D4E1}{0052}\pdfglyphtounicode{u1D4E2}{0053}
\pdfglyphtounicode{u1D4E3}{0054}\pdfglyphtounicode{u1D4E4}{0055}\pdfglyphtounicode{u1D4E5}{0056}\pdfglyphtounicode{u1D4E6}{0057}\pdfglyphtounicode{u1D4E7}{0058}\pdfglyphtounicode{u1D4E8}{0059}
\pdfglyphtounicode{u1D4E9}{005A}\pdfglyphtounicode{u1D4EA}{0061}\pdfglyphtounicode{u1D4EB}{0062}\pdfglyphtounicode{u1D4EC}{0063}\pdfglyphtounicode{u1D4ED}{0064}\pdfglyphtounicode{u1D4EE}{0065}
\pdfglyphtounicode{u1D4EF}{0066}\pdfglyphtounicode{u1D4F0}{0067}\pdfglyphtounicode{u1D4F1}{0068}\pdfglyphtounicode{u1D4F2}{0069}\pdfglyphtounicode{u1D4F3}{006A}\pdfglyphtounicode{u1D4F4}{006B}
\pdfglyphtounicode{u1D4F5}{006C}\pdfglyphtounicode{u1D4F6}{006D}\pdfglyphtounicode{u1D4F7}{006E}\pdfglyphtounicode{u1D4F8}{006F}\pdfglyphtounicode{u1D4F9}{0070}\pdfglyphtounicode{u1D4FA}{0071}
\pdfglyphtounicode{u1D4FB}{0072}\pdfglyphtounicode{u1D4FC}{0073}\pdfglyphtounicode{u1D4FD}{0074}\pdfglyphtounicode{u1D4FE}{0075}\pdfglyphtounicode{u1D4FF}{0076}\pdfglyphtounicode{u1D500}{0077}
\pdfglyphtounicode{u1D501}{0078}\pdfglyphtounicode{u1D502}{0079}\pdfglyphtounicode{u1D503}{007A}\pdfglyphtounicode{u1D504}{0041}\pdfglyphtounicode{u1D505}{0042}\pdfglyphtounicode{u1D506}{0043}
\pdfglyphtounicode{u1D507}{0044}\pdfglyphtounicode{u1D508}{0045}\pdfglyphtounicode{u1D509}{0046}\pdfglyphtounicode{u1D50A}{0047}\pdfglyphtounicode{u1D50B}{0048}\pdfglyphtounicode{u1D50C}{0049}
\pdfglyphtounicode{u1D50D}{004A}\pdfglyphtounicode{u1D50E}{004B}\pdfglyphtounicode{u1D50F}{004C}\pdfglyphtounicode{u1D510}{004D}\pdfglyphtounicode{u1D511}{004E}\pdfglyphtounicode{u1D512}{004F}
\pdfglyphtounicode{u1D513}{0050}\pdfglyphtounicode{u1D514}{0051}\pdfglyphtounicode{u1D515}{0052}\pdfglyphtounicode{u1D516}{0053}\pdfglyphtounicode{u1D517}{0054}\pdfglyphtounicode{u1D518}{0055}
\pdfglyphtounicode{u1D519}{0056}\pdfglyphtounicode{u1D51A}{0057}\pdfglyphtounicode{u1D51B}{0058}\pdfglyphtounicode{u1D51C}{0059}\pdfglyphtounicode{u1D51D}{005A}\pdfglyphtounicode{u1D51E}{0061}
\pdfglyphtounicode{u1D51F}{0062}\pdfglyphtounicode{u1D520}{0063}\pdfglyphtounicode{u1D521}{0064}\pdfglyphtounicode{u1D522}{0065}\pdfglyphtounicode{u1D523}{0066}\pdfglyphtounicode{u1D524}{0067}
\pdfglyphtounicode{u1D525}{0068}\pdfglyphtounicode{u1D526}{0069}\pdfglyphtounicode{u1D527}{006A}\pdfglyphtounicode{u1D528}{006B}\pdfglyphtounicode{u1D529}{006C}\pdfglyphtounicode{u1D52A}{006D}
\pdfglyphtounicode{u1D52B}{006E}\pdfglyphtounicode{u1D52C}{006F}\pdfglyphtounicode{u1D52D}{0070}\pdfglyphtounicode{u1D52E}{0071}\pdfglyphtounicode{u1D52F}{0072}\pdfglyphtounicode{u1D530}{0073}
\pdfglyphtounicode{u1D531}{0074}\pdfglyphtounicode{u1D532}{0075}\pdfglyphtounicode{u1D533}{0076}\pdfglyphtounicode{u1D534}{0077}\pdfglyphtounicode{u1D535}{0078}\pdfglyphtounicode{u1D536}{0079}
\pdfglyphtounicode{u1D537}{007A}\pdfglyphtounicode{u1D538}{0041}\pdfglyphtounicode{u1D539}{0042}\pdfglyphtounicode{u1D53A}{0043}\pdfglyphtounicode{u1D53B}{0044}\pdfglyphtounicode{u1D53C}{0045}
\pdfglyphtounicode{u1D53D}{0046}\pdfglyphtounicode{u1D53E}{0047}\pdfglyphtounicode{u1D53F}{0048}\pdfglyphtounicode{u1D540}{0049}\pdfglyphtounicode{u1D541}{004A}\pdfglyphtounicode{u1D542}{004B}
\pdfglyphtounicode{u1D543}{004C}\pdfglyphtounicode{u1D544}{004D}\pdfglyphtounicode{u1D545}{004E}\pdfglyphtounicode{u1D546}{004F}\pdfglyphtounicode{u1D547}{0050}\pdfglyphtounicode{u1D548}{0051}
\pdfglyphtounicode{u1D549}{0052}\pdfglyphtounicode{u1D54A}{0053}\pdfglyphtounicode{u1D54B}{0054}\pdfglyphtounicode{u1D54C}{0055}\pdfglyphtounicode{u1D54D}{0056}\pdfglyphtounicode{u1D54E}{0057}
\pdfglyphtounicode{u1D54F}{0058}\pdfglyphtounicode{u1D550}{0059}\pdfglyphtounicode{u1D551}{005A}\pdfglyphtounicode{u1D552}{0061}\pdfglyphtounicode{u1D553}{0062}\pdfglyphtounicode{u1D554}{0063}
\pdfglyphtounicode{u1D555}{0064}\pdfglyphtounicode{u1D556}{0065}\pdfglyphtounicode{u1D557}{0066}\pdfglyphtounicode{u1D558}{0067}\pdfglyphtounicode{u1D559}{0068}\pdfglyphtounicode{u1D55A}{0069}
\pdfglyphtounicode{u1D55B}{006A}\pdfglyphtounicode{u1D55C}{006B}\pdfglyphtounicode{u1D55D}{006C}\pdfglyphtounicode{u1D55E}{006D}\pdfglyphtounicode{u1D55F}{006E}\pdfglyphtounicode{u1D560}{006F}
\pdfglyphtounicode{u1D561}{0070}\pdfglyphtounicode{u1D562}{0071}\pdfglyphtounicode{u1D563}{0072}\pdfglyphtounicode{u1D564}{0073}\pdfglyphtounicode{u1D565}{0074}\pdfglyphtounicode{u1D566}{0075}
\pdfglyphtounicode{u1D567}{0076}\pdfglyphtounicode{u1D568}{0077}\pdfglyphtounicode{u1D569}{0078}\pdfglyphtounicode{u1D56A}{0079}\pdfglyphtounicode{u1D56B}{007A}\pdfglyphtounicode{u1D56C}{0041}
\pdfglyphtounicode{u1D56D}{0042}\pdfglyphtounicode{u1D56E}{0043}\pdfglyphtounicode{u1D56F}{0044}\pdfglyphtounicode{u1D570}{0045}\pdfglyphtounicode{u1D571}{0046}\pdfglyphtounicode{u1D572}{0047}
\pdfglyphtounicode{u1D573}{0048}\pdfglyphtounicode{u1D574}{0049}\pdfglyphtounicode{u1D575}{004A}\pdfglyphtounicode{u1D576}{004B}\pdfglyphtounicode{u1D577}{004C}\pdfglyphtounicode{u1D578}{004D}
\pdfglyphtounicode{u1D579}{004E}\pdfglyphtounicode{u1D57A}{004F}\pdfglyphtounicode{u1D57B}{0050}\pdfglyphtounicode{u1D57C}{0051}\pdfglyphtounicode{u1D57D}{0052}\pdfglyphtounicode{u1D57E}{0053}
\pdfglyphtounicode{u1D57F}{0054}\pdfglyphtounicode{u1D580}{0055}\pdfglyphtounicode{u1D581}{0056}\pdfglyphtounicode{u1D582}{0057}\pdfglyphtounicode{u1D583}{0058}\pdfglyphtounicode{u1D584}{0059}
\pdfglyphtounicode{u1D585}{005A}\pdfglyphtounicode{u1D586}{0061}\pdfglyphtounicode{u1D587}{0062}\pdfglyphtounicode{u1D588}{0063}\pdfglyphtounicode{u1D589}{0064}\pdfglyphtounicode{u1D58A}{0065}
\pdfglyphtounicode{u1D58B}{0066}\pdfglyphtounicode{u1D58C}{0067}\pdfglyphtounicode{u1D58D}{0068}\pdfglyphtounicode{u1D58E}{0069}\pdfglyphtounicode{u1D58F}{006A}\pdfglyphtounicode{u1D590}{006B}
\pdfglyphtounicode{u1D591}{006C}\pdfglyphtounicode{u1D592}{006D}\pdfglyphtounicode{u1D593}{006E}\pdfglyphtounicode{u1D594}{006F}\pdfglyphtounicode{u1D595}{0070}\pdfglyphtounicode{u1D596}{0071}
\pdfglyphtounicode{u1D597}{0072}\pdfglyphtounicode{u1D598}{0073}\pdfglyphtounicode{u1D599}{0074}\pdfglyphtounicode{u1D59A}{0075}\pdfglyphtounicode{u1D59B}{0076}\pdfglyphtounicode{u1D59C}{0077}
\pdfglyphtounicode{u1D59D}{0078}\pdfglyphtounicode{u1D59E}{0079}\pdfglyphtounicode{u1D59F}{007A}\pdfglyphtounicode{u1D5A0}{0041}\pdfglyphtounicode{u1D5A1}{0042}\pdfglyphtounicode{u1D5A2}{0043}
\pdfglyphtounicode{u1D5A3}{0044}\pdfglyphtounicode{u1D5A4}{0045}\pdfglyphtounicode{u1D5A5}{0046}\pdfglyphtounicode{u1D5A6}{0047}\pdfglyphtounicode{u1D5A7}{0048}\pdfglyphtounicode{u1D5A8}{0049}
\pdfglyphtounicode{u1D5A9}{004A}\pdfglyphtounicode{u1D5AA}{004B}\pdfglyphtounicode{u1D5AB}{004C}\pdfglyphtounicode{u1D5AC}{004D}\pdfglyphtounicode{u1D5AD}{004E}\pdfglyphtounicode{u1D5AE}{004F}
\pdfglyphtounicode{u1D5AF}{0050}\pdfglyphtounicode{u1D5B0}{0051}\pdfglyphtounicode{u1D5B1}{0052}\pdfglyphtounicode{u1D5B2}{0053}\pdfglyphtounicode{u1D5B3}{0054}\pdfglyphtounicode{u1D5B4}{0055}
\pdfglyphtounicode{u1D5B5}{0056}\pdfglyphtounicode{u1D5B6}{0057}\pdfglyphtounicode{u1D5B7}{0058}\pdfglyphtounicode{u1D5B8}{0059}\pdfglyphtounicode{u1D5B9}{005A}\pdfglyphtounicode{u1D5BA}{0061}
\pdfglyphtounicode{u1D5BB}{0062}\pdfglyphtounicode{u1D5BC}{0063}\pdfglyphtounicode{u1D5BD}{0064}\pdfglyphtounicode{u1D5BE}{0065}\pdfglyphtounicode{u1D5BF}{0066}\pdfglyphtounicode{u1D5C0}{0067}
\pdfglyphtounicode{u1D5C1}{0068}\pdfglyphtounicode{u1D5C2}{0069}\pdfglyphtounicode{u1D5C3}{006A}\pdfglyphtounicode{u1D5C4}{006B}\pdfglyphtounicode{u1D5C5}{006C}\pdfglyphtounicode{u1D5C6}{006D}
\pdfglyphtounicode{u1D5C7}{006E}\pdfglyphtounicode{u1D5C8}{006F}\pdfglyphtounicode{u1D5C9}{0070}\pdfglyphtounicode{u1D5CA}{0071}\pdfglyphtounicode{u1D5CB}{0072}\pdfglyphtounicode{u1D5CC}{0073}
\pdfglyphtounicode{u1D5CD}{0074}\pdfglyphtounicode{u1D5CE}{0075}\pdfglyphtounicode{u1D5CF}{0076}\pdfglyphtounicode{u1D5D0}{0077}\pdfglyphtounicode{u1D5D1}{0078}\pdfglyphtounicode{u1D5D2}{0079}
\pdfglyphtounicode{u1D5D3}{007A}\pdfglyphtounicode{u1D5D4}{0041}\pdfglyphtounicode{u1D5D5}{0042}\pdfglyphtounicode{u1D5D6}{0043}\pdfglyphtounicode{u1D5D7}{0044}\pdfglyphtounicode{u1D5D8}{0045}
\pdfglyphtounicode{u1D5D9}{0046}\pdfglyphtounicode{u1D5DA}{0047}\pdfglyphtounicode{u1D5DB}{0048}\pdfglyphtounicode{u1D5DC}{0049}\pdfglyphtounicode{u1D5DD}{004A}\pdfglyphtounicode{u1D5DE}{004B}
\pdfglyphtounicode{u1D5DF}{004C}\pdfglyphtounicode{u1D5E0}{004D}\pdfglyphtounicode{u1D5E1}{004E}\pdfglyphtounicode{u1D5E2}{004F}\pdfglyphtounicode{u1D5E3}{0050}\pdfglyphtounicode{u1D5E4}{0051}
\pdfglyphtounicode{u1D5E5}{0052}\pdfglyphtounicode{u1D5E6}{0053}\pdfglyphtounicode{u1D5E7}{0054}\pdfglyphtounicode{u1D5E8}{0055}\pdfglyphtounicode{u1D5E9}{0056}\pdfglyphtounicode{u1D5EA}{0057}
\pdfglyphtounicode{u1D5EB}{0058}\pdfglyphtounicode{u1D5EC}{0059}\pdfglyphtounicode{u1D5ED}{005A}\pdfglyphtounicode{u1D5EE}{0061}\pdfglyphtounicode{u1D5EF}{0062}\pdfglyphtounicode{u1D5F0}{0063}
\pdfglyphtounicode{u1D5F1}{0064}\pdfglyphtounicode{u1D5F2}{0065}\pdfglyphtounicode{u1D5F3}{0066}\pdfglyphtounicode{u1D5F4}{0067}\pdfglyphtounicode{u1D5F5}{0068}\pdfglyphtounicode{u1D5F6}{0069}
\pdfglyphtounicode{u1D5F7}{006A}\pdfglyphtounicode{u1D5F8}{006B}\pdfglyphtounicode{u1D5F9}{006C}\pdfglyphtounicode{u1D5FA}{006D}\pdfglyphtounicode{u1D5FB}{006E}\pdfglyphtounicode{u1D5FC}{006F}
\pdfglyphtounicode{u1D5FD}{0070}\pdfglyphtounicode{u1D5FE}{0071}\pdfglyphtounicode{u1D5FF}{0072}\pdfglyphtounicode{u1D600}{0073}\pdfglyphtounicode{u1D601}{0074}\pdfglyphtounicode{u1D602}{0075}
\pdfglyphtounicode{u1D603}{0076}\pdfglyphtounicode{u1D604}{0077}\pdfglyphtounicode{u1D605}{0078}\pdfglyphtounicode{u1D606}{0079}\pdfglyphtounicode{u1D607}{007A}\pdfglyphtounicode{u1D608}{0041}
\pdfglyphtounicode{u1D609}{0042}\pdfglyphtounicode{u1D60A}{0043}\pdfglyphtounicode{u1D60B}{0044}\pdfglyphtounicode{u1D60C}{0045}\pdfglyphtounicode{u1D60D}{0046}\pdfglyphtounicode{u1D60E}{0047}
\pdfglyphtounicode{u1D60F}{0048}\pdfglyphtounicode{u1D610}{0049}\pdfglyphtounicode{u1D611}{004A}\pdfglyphtounicode{u1D612}{004B}\pdfglyphtounicode{u1D613}{004C}\pdfglyphtounicode{u1D614}{004D}
\pdfglyphtounicode{u1D615}{004E}\pdfglyphtounicode{u1D616}{004F}\pdfglyphtounicode{u1D617}{0050}\pdfglyphtounicode{u1D618}{0051}\pdfglyphtounicode{u1D619}{0052}\pdfglyphtounicode{u1D61A}{0053}
\pdfglyphtounicode{u1D61B}{0054}\pdfglyphtounicode{u1D61C}{0055}\pdfglyphtounicode{u1D61D}{0056}\pdfglyphtounicode{u1D61E}{0057}\pdfglyphtounicode{u1D61F}{0058}\pdfglyphtounicode{u1D620}{0059}
\pdfglyphtounicode{u1D621}{005A}\pdfglyphtounicode{u1D622}{0061}\pdfglyphtounicode{u1D623}{0062}\pdfglyphtounicode{u1D624}{0063}\pdfglyphtounicode{u1D625}{0064}\pdfglyphtounicode{u1D626}{0065}
\pdfglyphtounicode{u1D627}{0066}\pdfglyphtounicode{u1D628}{0067}\pdfglyphtounicode{u1D629}{0068}\pdfglyphtounicode{u1D62A}{0069}\pdfglyphtounicode{u1D62B}{006A}\pdfglyphtounicode{u1D62C}{006B}
\pdfglyphtounicode{u1D62D}{006C}\pdfglyphtounicode{u1D62E}{006D}\pdfglyphtounicode{u1D62F}{006E}\pdfglyphtounicode{u1D630}{006F}\pdfglyphtounicode{u1D631}{0070}\pdfglyphtounicode{u1D632}{0071}
\pdfglyphtounicode{u1D633}{0072}\pdfglyphtounicode{u1D634}{0073}\pdfglyphtounicode{u1D635}{0074}\pdfglyphtounicode{u1D636}{0075}\pdfglyphtounicode{u1D637}{0076}\pdfglyphtounicode{u1D638}{0077}
\pdfglyphtounicode{u1D639}{0078}\pdfglyphtounicode{u1D63A}{0079}\pdfglyphtounicode{u1D63B}{007A}\pdfglyphtounicode{u1D63C}{0041}\pdfglyphtounicode{u1D63D}{0042}\pdfglyphtounicode{u1D63E}{0043}
\pdfglyphtounicode{u1D63F}{0044}\pdfglyphtounicode{u1D640}{0045}\pdfglyphtounicode{u1D641}{0046}\pdfglyphtounicode{u1D642}{0047}\pdfglyphtounicode{u1D643}{0048}\pdfglyphtounicode{u1D644}{0049}
\pdfglyphtounicode{u1D645}{004A}\pdfglyphtounicode{u1D646}{004B}\pdfglyphtounicode{u1D647}{004C}\pdfglyphtounicode{u1D648}{004D}\pdfglyphtounicode{u1D649}{004E}\pdfglyphtounicode{u1D64A}{004F}
\pdfglyphtounicode{u1D64B}{0050}\pdfglyphtounicode{u1D64C}{0051}\pdfglyphtounicode{u1D64D}{0052}\pdfglyphtounicode{u1D64E}{0053}\pdfglyphtounicode{u1D64F}{0054}\pdfglyphtounicode{u1D650}{0055}
\pdfglyphtounicode{u1D651}{0056}\pdfglyphtounicode{u1D652}{0057}\pdfglyphtounicode{u1D653}{0058}\pdfglyphtounicode{u1D654}{0059}\pdfglyphtounicode{u1D655}{005A}\pdfglyphtounicode{u1D656}{0061}
\pdfglyphtounicode{u1D657}{0062}\pdfglyphtounicode{u1D658}{0063}\pdfglyphtounicode{u1D659}{0064}\pdfglyphtounicode{u1D65A}{0065}\pdfglyphtounicode{u1D65B}{0066}\pdfglyphtounicode{u1D65C}{0067}
\pdfglyphtounicode{u1D65D}{0068}\pdfglyphtounicode{u1D65E}{0069}\pdfglyphtounicode{u1D65F}{006A}\pdfglyphtounicode{u1D660}{006B}\pdfglyphtounicode{u1D661}{006C}\pdfglyphtounicode{u1D662}{006D}
\pdfglyphtounicode{u1D663}{006E}\pdfglyphtounicode{u1D664}{006F}\pdfglyphtounicode{u1D665}{0070}\pdfglyphtounicode{u1D666}{0071}\pdfglyphtounicode{u1D667}{0072}\pdfglyphtounicode{u1D668}{0073}
\pdfglyphtounicode{u1D669}{0074}\pdfglyphtounicode{u1D66A}{0075}\pdfglyphtounicode{u1D66B}{0076}\pdfglyphtounicode{u1D66C}{0077}\pdfglyphtounicode{u1D66D}{0078}\pdfglyphtounicode{u1D66E}{0079}
\pdfglyphtounicode{u1D66F}{007A}\pdfglyphtounicode{u1D670}{0041}\pdfglyphtounicode{u1D671}{0042}\pdfglyphtounicode{u1D672}{0043}\pdfglyphtounicode{u1D673}{0044}\pdfglyphtounicode{u1D674}{0045}
\pdfglyphtounicode{u1D675}{0046}\pdfglyphtounicode{u1D676}{0047}\pdfglyphtounicode{u1D677}{0048}\pdfglyphtounicode{u1D678}{0049}\pdfglyphtounicode{u1D679}{004A}\pdfglyphtounicode{u1D67A}{004B}
\pdfglyphtounicode{u1D67B}{004C}\pdfglyphtounicode{u1D67C}{004D}\pdfglyphtounicode{u1D67D}{004E}\pdfglyphtounicode{u1D67E}{004F}\pdfglyphtounicode{u1D67F}{0050}\pdfglyphtounicode{u1D680}{0051}
\pdfglyphtounicode{u1D681}{0052}\pdfglyphtounicode{u1D682}{0053}\pdfglyphtounicode{u1D683}{0054}\pdfglyphtounicode{u1D684}{0055}\pdfglyphtounicode{u1D685}{0056}\pdfglyphtounicode{u1D686}{0057}
\pdfglyphtounicode{u1D687}{0058}\pdfglyphtounicode{u1D688}{0059}\pdfglyphtounicode{u1D689}{005A}\pdfglyphtounicode{u1D68A}{0061}\pdfglyphtounicode{u1D68B}{0062}\pdfglyphtounicode{u1D68C}{0063}
\pdfglyphtounicode{u1D68D}{0064}\pdfglyphtounicode{u1D68E}{0065}\pdfglyphtounicode{u1D68F}{0066}\pdfglyphtounicode{u1D690}{0067}\pdfglyphtounicode{u1D691}{0068}\pdfglyphtounicode{u1D692}{0069}
\pdfglyphtounicode{u1D693}{006A}\pdfglyphtounicode{u1D694}{006B}\pdfglyphtounicode{u1D695}{006C}\pdfglyphtounicode{u1D696}{006D}\pdfglyphtounicode{u1D697}{006E}\pdfglyphtounicode{u1D698}{006F}
\pdfglyphtounicode{u1D699}{0070}\pdfglyphtounicode{u1D69A}{0071}\pdfglyphtounicode{u1D69B}{0072}\pdfglyphtounicode{u1D69C}{0073}\pdfglyphtounicode{u1D69D}{0074}\pdfglyphtounicode{u1D69E}{0075}
\pdfglyphtounicode{u1D69F}{0076}\pdfglyphtounicode{u1D6A0}{0077}\pdfglyphtounicode{u1D6A1}{0078}\pdfglyphtounicode{u1D6A2}{0079}\pdfglyphtounicode{u1D6A3}{007A}\pdfglyphtounicode{u1D6A4}{0131}
\pdfglyphtounicode{u1D6A5}{0237}\pdfglyphtounicode{u1D6A8}{0391}\pdfglyphtounicode{u1D6A9}{0392}\pdfglyphtounicode{u1D6AA}{0393}\pdfglyphtounicode{u1D6AB}{0394}\pdfglyphtounicode{u1D6AC}{0395}
\pdfglyphtounicode{u1D6AD}{0396}\pdfglyphtounicode{u1D6AE}{0397}\pdfglyphtounicode{u1D6AF}{0398}\pdfglyphtounicode{u1D6B0}{0399}\pdfglyphtounicode{u1D6B1}{039A}\pdfglyphtounicode{u1D6B2}{039B}
\pdfglyphtounicode{u1D6B3}{039C}\pdfglyphtounicode{u1D6B4}{039D}\pdfglyphtounicode{u1D6B5}{039E}\pdfglyphtounicode{u1D6B6}{039F}\pdfglyphtounicode{u1D6B7}{03A0}\pdfglyphtounicode{u1D6B8}{03A1}
\pdfglyphtounicode{u1D6B9}{03F4}\pdfglyphtounicode{u1D6BA}{03A3}\pdfglyphtounicode{u1D6BB}{03A4}\pdfglyphtounicode{u1D6BC}{03A5}\pdfglyphtounicode{u1D6BD}{03A6}\pdfglyphtounicode{u1D6BE}{03A7}
\pdfglyphtounicode{u1D6BF}{03A8}\pdfglyphtounicode{u1D6C0}{03A9}\pdfglyphtounicode{u1D6C1}{2207}\pdfglyphtounicode{u1D6C2}{03B1}\pdfglyphtounicode{u1D6C3}{03B2}\pdfglyphtounicode{u1D6C4}{03B3}
\pdfglyphtounicode{u1D6C5}{03B4}\pdfglyphtounicode{u1D6C6}{03B5}\pdfglyphtounicode{u1D6C7}{03B6}\pdfglyphtounicode{u1D6C8}{03B7}\pdfglyphtounicode{u1D6C9}{03B8}\pdfglyphtounicode{u1D6CA}{03B9}
\pdfglyphtounicode{u1D6CB}{03BA}\pdfglyphtounicode{u1D6CC}{03BB}\pdfglyphtounicode{u1D6CD}{03BC}\pdfglyphtounicode{u1D6CE}{03BD}\pdfglyphtounicode{u1D6CF}{03BE}\pdfglyphtounicode{u1D6D0}{03BF}
\pdfglyphtounicode{u1D6D1}{03C0}\pdfglyphtounicode{u1D6D2}{03C1}\pdfglyphtounicode{u1D6D3}{03C2}\pdfglyphtounicode{u1D6D4}{03C3}\pdfglyphtounicode{u1D6D5}{03C4}\pdfglyphtounicode{u1D6D6}{03C5}
\pdfglyphtounicode{u1D6D7}{03C6}\pdfglyphtounicode{u1D6D8}{03C7}\pdfglyphtounicode{u1D6D9}{03C8}\pdfglyphtounicode{u1D6DA}{03C9}\pdfglyphtounicode{u1D6DB}{2202}\pdfglyphtounicode{u1D6DC}{03F5}
\pdfglyphtounicode{u1D6DD}{03D1}\pdfglyphtounicode{u1D6DE}{03F0}\pdfglyphtounicode{u1D6DF}{03D5}\pdfglyphtounicode{u1D6E0}{03F1}\pdfglyphtounicode{u1D6E1}{03D6}\pdfglyphtounicode{u1D6E2}{0391}
\pdfglyphtounicode{u1D6E3}{0392}\pdfglyphtounicode{u1D6E4}{0393}\pdfglyphtounicode{u1D6E5}{0394}\pdfglyphtounicode{u1D6E6}{0395}\pdfglyphtounicode{u1D6E7}{0396}\pdfglyphtounicode{u1D6E8}{0397}
\pdfglyphtounicode{u1D6E9}{0398}\pdfglyphtounicode{u1D6EA}{0399}\pdfglyphtounicode{u1D6EB}{039A}\pdfglyphtounicode{u1D6EC}{039B}\pdfglyphtounicode{u1D6ED}{039C}\pdfglyphtounicode{u1D6EE}{039D}
\pdfglyphtounicode{u1D6EF}{039E}\pdfglyphtounicode{u1D6F0}{039F}\pdfglyphtounicode{u1D6F1}{03A0}\pdfglyphtounicode{u1D6F2}{03A1}\pdfglyphtounicode{u1D6F3}{03F4}\pdfglyphtounicode{u1D6F4}{03A3}
\pdfglyphtounicode{u1D6F5}{03A4}\pdfglyphtounicode{u1D6F6}{03A5}\pdfglyphtounicode{u1D6F7}{03A6}\pdfglyphtounicode{u1D6F8}{03A7}\pdfglyphtounicode{u1D6F9}{03A8}\pdfglyphtounicode{u1D6FA}{03A9}
\pdfglyphtounicode{u1D6FB}{2207}\pdfglyphtounicode{u1D6FC}{03B1}\pdfglyphtounicode{u1D6FD}{03B2}\pdfglyphtounicode{u1D6FE}{03B3}\pdfglyphtounicode{u1D6FF}{03B4}\pdfglyphtounicode{u1D700}{03B5}
\pdfglyphtounicode{u1D701}{03B6}\pdfglyphtounicode{u1D702}{03B7}\pdfglyphtounicode{u1D703}{03B8}\pdfglyphtounicode{u1D704}{03B9}\pdfglyphtounicode{u1D705}{03BA}\pdfglyphtounicode{u1D706}{03BB}
\pdfglyphtounicode{u1D707}{03BC}\pdfglyphtounicode{u1D708}{03BD}\pdfglyphtounicode{u1D709}{03BE}\pdfglyphtounicode{u1D70A}{03BF}\pdfglyphtounicode{u1D70B}{03C0}\pdfglyphtounicode{u1D70C}{03C1}
\pdfglyphtounicode{u1D70D}{03C2}\pdfglyphtounicode{u1D70E}{03C3}\pdfglyphtounicode{u1D70F}{03C4}\pdfglyphtounicode{u1D710}{03C5}\pdfglyphtounicode{u1D711}{03C6}\pdfglyphtounicode{u1D712}{03C7}
\pdfglyphtounicode{u1D713}{03C8}\pdfglyphtounicode{u1D714}{03C9}\pdfglyphtounicode{u1D715}{2202}\pdfglyphtounicode{u1D716}{03F5}\pdfglyphtounicode{u1D717}{03D1}\pdfglyphtounicode{u1D718}{03F0}
\pdfglyphtounicode{u1D719}{03D5}\pdfglyphtounicode{u1D71A}{03F1}\pdfglyphtounicode{u1D71B}{03D6}\pdfglyphtounicode{u1D71C}{0391}\pdfglyphtounicode{u1D71D}{0392}\pdfglyphtounicode{u1D71E}{0393}
\pdfglyphtounicode{u1D71F}{0394}\pdfglyphtounicode{u1D720}{0395}\pdfglyphtounicode{u1D721}{0396}\pdfglyphtounicode{u1D722}{0397}\pdfglyphtounicode{u1D723}{0398}\pdfglyphtounicode{u1D724}{0399}
\pdfglyphtounicode{u1D725}{039A}\pdfglyphtounicode{u1D726}{039B}\pdfglyphtounicode{u1D727}{039C}\pdfglyphtounicode{u1D728}{039D}\pdfglyphtounicode{u1D729}{039E}\pdfglyphtounicode{u1D72A}{039F}
\pdfglyphtounicode{u1D72B}{03A0}\pdfglyphtounicode{u1D72C}{03A1}\pdfglyphtounicode{u1D72D}{03F4}\pdfglyphtounicode{u1D72E}{03A3}\pdfglyphtounicode{u1D72F}{03A4}\pdfglyphtounicode{u1D730}{03A5}
\pdfglyphtounicode{u1D731}{03A6}\pdfglyphtounicode{u1D732}{03A7}\pdfglyphtounicode{u1D733}{03A8}\pdfglyphtounicode{u1D734}{03A9}\pdfglyphtounicode{u1D735}{2207}\pdfglyphtounicode{u1D736}{03B1}
\pdfglyphtounicode{u1D737}{03B2}\pdfglyphtounicode{u1D738}{03B3}\pdfglyphtounicode{u1D739}{03B4}\pdfglyphtounicode{u1D73A}{03B5}\pdfglyphtounicode{u1D73B}{03B6}\pdfglyphtounicode{u1D73C}{03B7}
\pdfglyphtounicode{u1D73D}{03B8}\pdfglyphtounicode{u1D73E}{03B9}\pdfglyphtounicode{u1D73F}{03BA}\pdfglyphtounicode{u1D740}{03BB}\pdfglyphtounicode{u1D741}{03BC}\pdfglyphtounicode{u1D742}{03BD}
\pdfglyphtounicode{u1D743}{03BE}\pdfglyphtounicode{u1D744}{03BF}\pdfglyphtounicode{u1D745}{03C0}\pdfglyphtounicode{u1D746}{03C1}\pdfglyphtounicode{u1D747}{03C2}\pdfglyphtounicode{u1D748}{03C3}
\pdfglyphtounicode{u1D749}{03C4}\pdfglyphtounicode{u1D74A}{03C5}\pdfglyphtounicode{u1D74B}{03C6}\pdfglyphtounicode{u1D74C}{03C7}\pdfglyphtounicode{u1D74D}{03C8}\pdfglyphtounicode{u1D74E}{03C9}
\pdfglyphtounicode{u1D74F}{2202}\pdfglyphtounicode{u1D750}{03F5}\pdfglyphtounicode{u1D751}{03D1}\pdfglyphtounicode{u1D752}{03F0}\pdfglyphtounicode{u1D753}{03D5}\pdfglyphtounicode{u1D754}{03F1}
\pdfglyphtounicode{u1D755}{03D6}\pdfglyphtounicode{u1D756}{0391}\pdfglyphtounicode{u1D757}{0392}\pdfglyphtounicode{u1D758}{0393}\pdfglyphtounicode{u1D759}{0394}\pdfglyphtounicode{u1D75A}{0395}
\pdfglyphtounicode{u1D75B}{0396}\pdfglyphtounicode{u1D75C}{0397}\pdfglyphtounicode{u1D75D}{0398}\pdfglyphtounicode{u1D75E}{0399}\pdfglyphtounicode{u1D75F}{039A}\pdfglyphtounicode{u1D760}{039B}
\pdfglyphtounicode{u1D761}{039C}\pdfglyphtounicode{u1D762}{039D}\pdfglyphtounicode{u1D763}{039E}\pdfglyphtounicode{u1D764}{039F}\pdfglyphtounicode{u1D765}{03A0}\pdfglyphtounicode{u1D766}{03A1}
\pdfglyphtounicode{u1D767}{03F4}\pdfglyphtounicode{u1D768}{03A3}\pdfglyphtounicode{u1D769}{03A4}\pdfglyphtounicode{u1D76A}{03A5}\pdfglyphtounicode{u1D76B}{03A6}\pdfglyphtounicode{u1D76C}{03A7}
\pdfglyphtounicode{u1D76D}{03A8}\pdfglyphtounicode{u1D76E}{03A9}\pdfglyphtounicode{u1D76F}{2207}\pdfglyphtounicode{u1D770}{03B1}\pdfglyphtounicode{u1D771}{03B2}\pdfglyphtounicode{u1D772}{03B3}
\pdfglyphtounicode{u1D773}{03B4}\pdfglyphtounicode{u1D774}{03B5}\pdfglyphtounicode{u1D775}{03B6}\pdfglyphtounicode{u1D776}{03B7}\pdfglyphtounicode{u1D777}{03B8}\pdfglyphtounicode{u1D778}{03B9}
\pdfglyphtounicode{u1D779}{03BA}\pdfglyphtounicode{u1D77A}{03BB}\pdfglyphtounicode{u1D77B}{03BC}\pdfglyphtounicode{u1D77C}{03BD}\pdfglyphtounicode{u1D77D}{03BE}\pdfglyphtounicode{u1D77E}{03BF}
\pdfglyphtounicode{u1D77F}{03C0}\pdfglyphtounicode{u1D780}{03C1}\pdfglyphtounicode{u1D781}{03C2}\pdfglyphtounicode{u1D782}{03C3}\pdfglyphtounicode{u1D783}{03C4}\pdfglyphtounicode{u1D784}{03C5}
\pdfglyphtounicode{u1D785}{03C6}\pdfglyphtounicode{u1D786}{03C7}\pdfglyphtounicode{u1D787}{03C8}\pdfglyphtounicode{u1D788}{03C9}\pdfglyphtounicode{u1D789}{2202}\pdfglyphtounicode{u1D78A}{03F5}
\pdfglyphtounicode{u1D78B}{03D1}\pdfglyphtounicode{u1D78C}{03F0}\pdfglyphtounicode{u1D78D}{03D5}\pdfglyphtounicode{u1D78E}{03F1}\pdfglyphtounicode{u1D78F}{03D6}\pdfglyphtounicode{u1D790}{0391}
\pdfglyphtounicode{u1D791}{0392}\pdfglyphtounicode{u1D792}{0393}\pdfglyphtounicode{u1D793}{0394}\pdfglyphtounicode{u1D794}{0395}\pdfglyphtounicode{u1D795}{0396}\pdfglyphtounicode{u1D796}{0397}
\pdfglyphtounicode{u1D797}{0398}\pdfglyphtounicode{u1D798}{0399}\pdfglyphtounicode{u1D799}{039A}\pdfglyphtounicode{u1D79A}{039B}\pdfglyphtounicode{u1D79B}{039C}\pdfglyphtounicode{u1D79C}{039D}
\pdfglyphtounicode{u1D79D}{039E}\pdfglyphtounicode{u1D79E}{039F}\pdfglyphtounicode{u1D79F}{03A0}\pdfglyphtounicode{u1D7A0}{03A1}\pdfglyphtounicode{u1D7A1}{03F4}\pdfglyphtounicode{u1D7A2}{03A3}
\pdfglyphtounicode{u1D7A3}{03A4}\pdfglyphtounicode{u1D7A4}{03A5}\pdfglyphtounicode{u1D7A5}{03A6}\pdfglyphtounicode{u1D7A6}{03A7}\pdfglyphtounicode{u1D7A7}{03A8}\pdfglyphtounicode{u1D7A8}{03A9}
\pdfglyphtounicode{u1D7A9}{2207}\pdfglyphtounicode{u1D7AA}{03B1}\pdfglyphtounicode{u1D7AB}{03B2}\pdfglyphtounicode{u1D7AC}{03B3}\pdfglyphtounicode{u1D7AD}{03B4}\pdfglyphtounicode{u1D7AE}{03B5}
\pdfglyphtounicode{u1D7AF}{03B6}\pdfglyphtounicode{u1D7B0}{03B7}\pdfglyphtounicode{u1D7B1}{03B8}\pdfglyphtounicode{u1D7B2}{03B9}\pdfglyphtounicode{u1D7B3}{03BA}\pdfglyphtounicode{u1D7B4}{03BB}
\pdfglyphtounicode{u1D7B5}{03BC}\pdfglyphtounicode{u1D7B6}{03BD}\pdfglyphtounicode{u1D7B7}{03BE}\pdfglyphtounicode{u1D7B8}{03BF}\pdfglyphtounicode{u1D7B9}{03C0}\pdfglyphtounicode{u1D7BA}{03C1}
\pdfglyphtounicode{u1D7BB}{03C2}\pdfglyphtounicode{u1D7BC}{03C3}\pdfglyphtounicode{u1D7BD}{03C4}\pdfglyphtounicode{u1D7BE}{03C5}\pdfglyphtounicode{u1D7BF}{03C6}\pdfglyphtounicode{u1D7C0}{03C7}
\pdfglyphtounicode{u1D7C1}{03C8}\pdfglyphtounicode{u1D7C2}{03C9}\pdfglyphtounicode{u1D7C3}{2202}\pdfglyphtounicode{u1D7C4}{03F5}\pdfglyphtounicode{u1D7C5}{03D1}\pdfglyphtounicode{u1D7C6}{03F0}
\pdfglyphtounicode{u1D7C7}{03D5}\pdfglyphtounicode{u1D7C8}{03F1}\pdfglyphtounicode{u1D7C9}{03D6}\pdfglyphtounicode{u1D7CE}{0030}\pdfglyphtounicode{u1D7CF}{0031}\pdfglyphtounicode{u1D7D0}{0032}
\pdfglyphtounicode{u1D7D1}{0033}\pdfglyphtounicode{u1D7D2}{0034}\pdfglyphtounicode{u1D7D3}{0035}\pdfglyphtounicode{u1D7D4}{0036}\pdfglyphtounicode{u1D7D5}{0037}\pdfglyphtounicode{u1D7D6}{0038}
\pdfglyphtounicode{u1D7D7}{0039}\pdfglyphtounicode{u1D7D8}{0030}\pdfglyphtounicode{u1D7D9}{0031}\pdfglyphtounicode{u1D7DA}{0032}\pdfglyphtounicode{u1D7DB}{0033}\pdfglyphtounicode{u1D7DC}{0034}
\pdfglyphtounicode{u1D7DD}{0035}\pdfglyphtounicode{u1D7DE}{0036}\pdfglyphtounicode{u1D7DF}{0037}\pdfglyphtounicode{u1D7E0}{0038}\pdfglyphtounicode{u1D7E1}{0039}\pdfglyphtounicode{u1D7E2}{0030}
\pdfglyphtounicode{u1D7E3}{0031}\pdfglyphtounicode{u1D7E4}{0032}\pdfglyphtounicode{u1D7E5}{0033}\pdfglyphtounicode{u1D7E6}{0034}\pdfglyphtounicode{u1D7E7}{0035}\pdfglyphtounicode{u1D7E8}{0036}
\pdfglyphtounicode{u1D7E9}{0037}\pdfglyphtounicode{u1D7EA}{0038}\pdfglyphtounicode{u1D7EB}{0039}\pdfglyphtounicode{u1D7EC}{0030}\pdfglyphtounicode{u1D7ED}{0031}\pdfglyphtounicode{u1D7EE}{0032}
\pdfglyphtounicode{u1D7EF}{0033}\pdfglyphtounicode{u1D7F0}{0034}\pdfglyphtounicode{u1D7F1}{0035}\pdfglyphtounicode{u1D7F2}{0036}\pdfglyphtounicode{u1D7F3}{0037}\pdfglyphtounicode{u1D7F4}{0038}
\pdfglyphtounicode{u1D7F5}{0039}\pdfglyphtounicode{u1D7F6}{0030}\pdfglyphtounicode{u1D7F7}{0031}\pdfglyphtounicode{u1D7F8}{0032}\pdfglyphtounicode{u1D7F9}{0033}\pdfglyphtounicode{u1D7FA}{0034}
\pdfglyphtounicode{u1D7FB}{0035}\pdfglyphtounicode{u1D7FC}{0036}\pdfglyphtounicode{u1D7FD}{0037}\pdfglyphtounicode{u1D7FE}{0038}\pdfglyphtounicode{u1D7FF}{0039}
\fi

\newcommand{\method}{CoRef-DETR}
\newcommand{\methodhr}{CoRef-DETR-HR}
\newcommand{\sig}{$^{\ast}$}
\newcommand{\cmark}{\ding{51}}
\newcommand{\xmark}{\ding{55}}
\newcommand{\ind}{\mathbf{1}}
\newif\iforcidshow
\NewDocumentCommand{\orcidicon}{m}{\iforcidshow\,\orcidlink{#1}\fi}

\ExplSyntaxOn
\cs_set_eq:NN \__coref_stmauthors_orig: \stmauthors
\cs_set_protected:Npn \stmauthors { \orcidshowtrue \__coref_stmauthors_orig: \orcidshowfalse }
\cs_set_protected:Npn \printorcid { }
\cs_set:Npn \__first_footerline:
  { \group_begin: \small \sffamily \ifnum\theblind>0\relax \else \__short_authors: \fi \group_end: }
\ExplSyntaxOff
\makeatletter
\AtEndDocument{\clearpage\addtocounter{page}{-1}%
  \immediate\write\@auxout{\string\newlabel{coref-lastpage}{{}{\thepage}{}{}{}}}\addtocounter{page}{1}}
\makeatother
\AtBeginDocument{}
\hypersetup{pdfauthor={M. Fazri Nizar, Julian Supardi, Muhammad Naufal Rachmatullah}, pdfsubject={Underwater instance segmentation}, pdfkeywords={Underwater instance segmentation, Detection transformer, Global context, Boundary refinement, Category refinement, Efficient inference}}
\makeatletter
\AtEndDocument{\clearpage\immediate\write\@auxout{\string\csxdef{lastpage}{\the\numexpr\value{page}-1\relax}}}
\makeatother
\begin{document}
\let\WriteBookmarks\relax
\renewcommand{\topfraction}{0.92}
\renewcommand{\dbltopfraction}{0.92}
\renewcommand{\bottomfraction}{0.6}
\renewcommand{\textfraction}{0.06}
\renewcommand{\floatpagefraction}{0.66}
\renewcommand{\dblfloatpagefraction}{0.85}%
\setcounter{topnumber}{3}
\setcounter{dbltopnumber}{2}
\setcounter{totalnumber}{5}

\shorttitle{CoRef-DETR for underwater instance segmentation}
\shortauthors{M.F. Nizar et al.}

\title [mode = title]{\method{}: Decoupled Global Context and Selective Boundary-Category Refinement for Underwater Instance Segmentation}

\author[a]{M. Fazri Nizar\orcidicon{0009-0002-8330-8520}}
\ead{mfazrinizar@gmail.com}
\credit{Conceptualization, Methodology, Software, Investigation, Data curation, Formal analysis, Visualization, Writing -- original draft}

\author[a]{Julian Supardi\orcidicon{0000-0002-5836-9236}}
\cormark[1]
\ead{julian@unsri.ac.id}
\credit{Conceptualization, Supervision, Validation, Writing -- review \& editing}

\author[a]{Muhammad Naufal Rachmatullah\orcidicon{0000-0003-3553-3475}}
\ead{naufalrachmatullah@unsri.ac.id}
\credit{Supervision, Validation, Writing -- review \& editing}

\affiliation[a]{organization={Department of Informatics Engineering, Faculty of Computer Science, Universitas Sriwijaya},
            addressline={Jl. Raya Palembang--Prabumulih Km.~32},
            city={Indralaya, Ogan Ilir},
            postcode={30662},
            state={South Sumatra},
            country={Indonesia}}

\cortext[1]{Corresponding author.}

\begin{abstract}
Underwater instance segmentation supports marine surveys, aquaculture, and underwater robotics. Haze, colour casts, and low contrast blur object boundaries, and recent benchmarks distinguish up to 158 fine-grained categories. The strongest published methods mostly have 219 to 632 million parameters. This paper proposes \method{}, a compact detection transformer that targets three error sources of a real-time detector: context, boundaries, and categories. First, a decoupled global context (DGC) detector, with the base detector's 33.4\,M parameters, applies global self-attention to a 384-pixel token grid but supervises and decodes masks in a 576-pixel frame. Second, a posterior-conditioned boundary refiner with 47.7\,k parameters corrects mask logits on predicted boundaries, and a quality score on the corrected masks re-ranks the detections. Third, a crop-probe category branch on frozen DINOv2 features revises the categories of at most three masks per image, where a rule fixed on validation data chooses soft fusion or a confident-disagreement gate. On five protocols of UIIS, UIIS10K, USIS10K, and USIS16K, \method{} achieves state-of-the-art mask average precision (AP), exceeding the strongest published results by 2.3 to 5.2 points, and outperforms three reproduced recent methods on the same test images (on USIS16K, with its category branch). In half precision with TensorRT, it processes an image in a median of 34 to 40\,ms, JPEG decoding included, on a T4 GPU, 1.4 to 11 times faster than these methods. Ablations attribute most of DGC's gain to the larger frame; an error analysis locates the remaining error in crowds of small objects.
\end{abstract}


\maketitle
\hypersetup{pdfauthor={M. Fazri Nizar, Julian Supardi, Muhammad Naufal Rachmatullah}, pdfsubject={Underwater instance segmentation}, pdfkeywords={Underwater instance segmentation, Detection transformer, Global context, Boundary refinement, Category refinement, Efficient inference}}

\section{Introduction}
\label{sec:intro}

Instance segmentation of underwater imagery delineates and classifies every fish, diver, robot, coral, or wreck in an image. Ecological surveys, aquaculture monitoring, and the perception systems of underwater vehicles all depend on it. Dedicated benchmarks now cover generic categories (UIIS~\cite{lian2023watermask} and UIIS10K~\cite{li2025uwsam}), salient instances (USIS10K~\cite{lian2024usissam}), and a fine-grained taxonomy of 158 categories (USIS16K~\cite{hong2025usis16k}). Water absorbs and scatters light unevenly across wavelengths~\cite{akkaynak2018revised}, so these images suffer from haze, colour casts, low contrast, and blurred object boundaries. In addition, many marine species differ only in subtle texture and shape.

Most recent progress on these benchmarks comes from large pretrained models: segmenters built on the Segment Anything Model (SAM)~\cite{lian2024usissam,li2025uwsam,ma2026bipa}, DINOv2 backbones with underwater adaptation~\cite{chen2026diveseg}, and designs that add estimated depth~\cite{shang2026spdasam,zheng2026dcfnet}. The strongest published results on four of the five evaluation protocols (USIS10K is scored both class-agnostically and with seven categories) come from models with 219 to 632\,M parameters; on class-agnostic USIS10K, a 28\,M YOLO-based model reports the best result~\cite{kong2025aasnet} (Table~\ref{tab:sota}). Underwater vehicles and edge platforms rarely offer the memory or the latency budget such models need. A compact detector as accurate as these models on all four benchmarks would serve these settings better.

\begin{figure*}[pos=t]
\centering
\includegraphics[width=\textwidth]{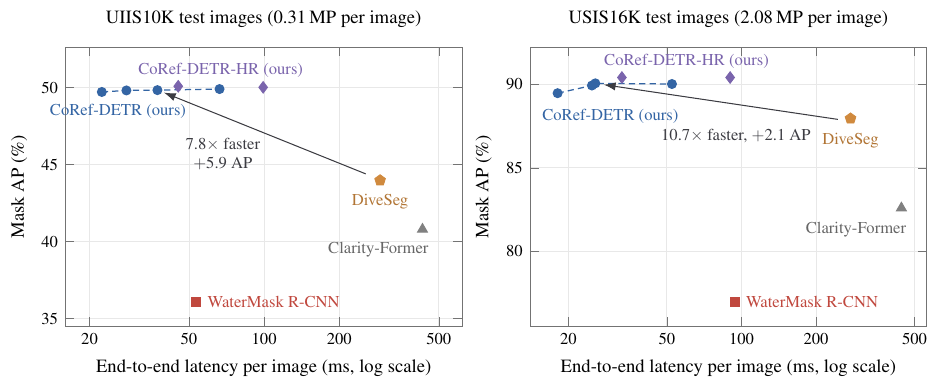}
\caption{Accuracy against end-to-end latency on an NVIDIA T4, every method running as a TensorRT pipeline on the same test images. Circles joined by a dashed line: operating points of \method{} (FP32 detector rightmost; leftward, an FP16 detector, a refiner limited to 32 patches per mask, and a DINOv2-S crop encoder); diamonds: \methodhr{} (FP16 left, FP32 right). Our AP includes CPC; a competing method's AP is its value in the controlled comparison (Section~\ref{sec:exp-same}). Arrows: speed-up and AP gain over DiveSeg, the strongest competing method. Latency is the median over 200 test images, from upload of the decoded image to the final outputs (Section~\ref{sec:exp-eff}).}
\label{fig:teaser}
\end{figure*}

This work starts from RF-DETR-Seg~\cite{robinson2025nasdetr}, a real-time detection transformer with a DINOv2 backbone, and examines where it still makes errors on underwater data. Three sources of error stand out, each with its own trade-off. The first is context. Reading the image at 576 instead of 384 pixels gives the largest gain we measured on the detector side, 1.51 points of mask average precision (AP), pooled over UIIS10K and USIS10K-A, but it multiplies the detector's FLOPs by 2.6 and its latency by 2.5. The second is the object boundary. The detector's mask posterior is uncertain where haze and low contrast blur the outline, and a small refiner that corrects this posterior along predicted boundaries recovers 0.3 AP on USIS16K to 1.4 on UIIS10K. The third is the category. On USIS16K, scoring the same predictions class-agnostically raises AP by 1.4 to 1.9 points, so much of the remaining error lies in category assignment. Fusing a crop classifier's category evidence raises validation AP on USIS16K by about 2 points, yet the same fusion lowers it by up to 2.3 points on the coarser taxonomies.

The \textbf{Co}ntext and \textbf{Ref}inement \textbf{De}tection \textbf{Tr}ansformer (\method{}) addresses the three sources with three stages, each changing a different output: a detector that computes global context on a compact token grid but is supervised and decodes masks in a larger frame, a boundary refiner paired with a quality score, and a category branch whose mode a rule fixed on validation data selects for each protocol.

Experiments on UIIS, UIIS10K, USIS10K, and USIS16K show that \method{} sets a new state of the art on all five evaluation protocols (Table~\ref{tab:sota}). With a detector of 33.4\,M parameters, it exceeds the best published mask AP on every protocol, by 2.25 to 5.20 points. When all methods are scored by one evaluator on the same test images, it also outperforms WaterMask R-CNN~\cite{lian2023watermask}, Clarity-Former~\cite{chen2026clarityformer}, and DiveSeg~\cite{chen2026diveseg} on every protocol (on USIS16K, with the category branch). With an FP16 detector, its TensorRT pipeline processes an image in a median of 26 to 37\,ms on a T4 GPU (34 to 40\,ms with JPEG decoding), 1.4 to 11 times faster than these methods (Fig.~\ref{fig:teaser}).

The main contributions of this paper are summarised as follows:
\begin{itemize}\setlength{\itemsep}{0pt}\setlength{\parsep}{0pt}\setlength{\parskip}{0pt}
\item We proposed \method{}, a compact framework for underwater instance segmentation that improved a real-time detection transformer in three stages, each aimed at one source of error: context, boundaries, and categories.
\item We designed a decoupled global context (DGC) detector that applied global self-attention to a compact 384-pixel token grid while supervising and decoding masks in a 576-pixel frame. Without adding parameters, it recovered 54\,\% of the gain of 576-pixel inputs for 19\,\% of the FLOPs they added. A factorial ablation on two protocols showed that the larger supervision and decoding frame carried most of this gain (0.52 of 0.81 AP), while global attention on the compact grid added a smaller share.
\item We introduced a lightweight posterior-conditioned boundary refiner (PBR) that corrected mask logits along predicted boundaries, and paired it with a quality score (Q) that re-ranked the detections by an IoU estimate computed on the corrected masks. Fitted to the predictions of three other detectors, PBR also improved them on four of the six detector--protocol pairs tested. Its gain depended more on the benchmark it was fitted on than on the detector, partly because the benchmarks differ in how densely their polygons trace object boundaries.
\item We developed a crop-probe category branch (CPC) that revised the categories of the top-ranked masks with a linear probe on frozen DINOv2 features while keeping the detector's own category hypotheses instead of overwriting them. A rule fixed on validation data chose between soft fusion and a gate that acted only on confident disagreements.
\item We showed that the three models trained with different random seeds for the evaluation could be averaged into the deployed model. The weight average added no inference cost, gave the largest post-training gain, and worked for a detection transformer because fine-tuning from one checkpoint kept the learned query slots aligned: each of the 100 queries was closest to the same query in the other two seed models (cosine 1.00, against 0.01 for any other query).
\item Experiments on five protocols of four underwater benchmarks, with one evaluator, one latency protocol, and paired bootstrap intervals, showed that \method{} achieved state-of-the-art accuracy and outperformed three reproduced recent methods on every protocol (on USIS16K, once the category branch was added) while running faster than them. Ablations examined each component, and an error analysis showed that crowding and object size accounted for the largest differences in accuracy across test images.
\end{itemize}
\section{Related work}
\label{sec:related}

\subsection{Underwater instance segmentation}

Benchmarks for underwater instance segmentation have grown in size and in the granularity of their labels. WaterMask~\cite{lian2023watermask} introduced UIIS with seven generic categories, together with a Mask R-CNN~\cite{he2017maskrcnn} variant that uses graph attention to recover details lost to degradation and a boundary-mask strategy to sharpen object boundaries. USIS-SAM~\cite{lian2024usissam} introduced USIS10K for salient instances, benchmarked salient-instance segmenters such as RDPNet~\cite{wu2021rdpnet} and OQTR~\cite{pei2023oqtr} on it, and adapted the Segment Anything Model~\cite{kirillov2023sam} to the task. UWSAM~\cite{li2025uwsam} contributed UIIS10K with ten categories and distilled a SAM teacher built on a huge vision transformer (ViT-H) into an efficient student, and USIS16K~\cite{hong2025usis16k} extended salient instance segmentation to 158 fine-grained categories. The strongest recent methods adapt large pretrained models: DiveSeg~\cite{chen2026diveseg} aligns DINOv2 features to underwater styles and prompts them with object priors, BiPA~\cite{ma2026bipa} adapts SAM with bilevel prompts, SPDA-SAM~\cite{shang2026spdasam} adds self-prompting and depth awareness to SAM, and ViT-UWA~\cite{jia2026vituwa} adapts a plain vision transformer to dense underwater prediction. Other methods use state-space backbones (UIS-Mamba~\cite{cong2025uismamba}), pseudo-depth fusion (DCFNet~\cite{zheng2026dcfnet}), Gaussian-mixture heatmap supervision (USIS-PGM~\cite{hong2026usispgm}), a boundary-aware decoder with environment-robust adapters (BARD-ERA~\cite{pan2025baris}), or transformers designed against degradation and geometric diversity (Clarity-Former~\cite{chen2026clarityformer}). AASNet~\cite{kong2025aasnet} extends a YOLOv9 detector with correlation attention and an adaptive focal loss. Related problems have also been studied: CAM-RCNN~\cite{yi2024camrcnn} adds coordinate-aware convolutions and group normalisation to Mask R-CNN for marine animals, WaterBox~\cite{wu2025waterbox} learns underwater instance segmentation from box labels, YOSAM2~\cite{li2026yosam2} combines a YOLO detector with SAM~2, and SU-YOLO~\cite{li2025suyolo} uses spiking neural networks for efficient underwater detection. Camouflaged instance segmentation~\cite{wang2026ucissam} is a closely related task, and Supplementary Section~S1 reviews foundation models for marine animals. Image enhancement as pre-processing has also been studied for segmentation on UIIS~\cite{lin2026dtiuie}; \method{} works on the raw images.

Most of these methods raise accuracy by adapting a larger backbone or foundation model. \method{} keeps a 33\,M-parameter detector, initialised from DINOv2-S, and adds two lightweight components of its own after it: a 47.7\,k-parameter refiner and, optionally, a linear probe on a frozen DINOv2-B (120\,M parameters in all). We compare it with three of these methods on the same test images, and with the remaining methods through the results their papers report on our protocols (Table~\ref{tab:sota}).

\subsection{Detection transformers, resolution, and context}

DETR~\cite{carion2020detr} cast detection as set prediction with learned queries. Deformable attention~\cite{zhu2021deformabledetr}, query denoising and mixed query selection~\cite{zhang2023dino}, and real-time designs~\cite{zhao2024rtdetr,chen2024lwdetr} made such detectors fast and accurate, and query-based models extend them to masks~\cite{fang2021queryinst,cheng2022mask2former,li2023maskdino}. RF-DETR~\cite{robinson2025nasdetr} pairs a DINOv2 backbone~\cite{oquab2023dinov2} with a lightweight detection transformer and searches resolution, patch size, attention windows, and other dimensions for real-time operating points; its segmentation variant is our base detector. Several designs compute global context only where it is cheap: plain-ViT backbones attend within windows and exchange information at a few global layers~\cite{li2022vitdet}, LW-DETR interleaves window and global attention~\cite{chen2024lwdetr}, RT-DETR restricts self-attention to its lowest-resolution feature map~\cite{zhao2024rtdetr}, and S-ViT computes self-attention on salient tokens only~\cite{zhou2026svit}. Lightweight detectors for driving and aerial scenes, such as SDF-YOLO~\cite{lu2026sdfyolo} and MEF-DETR~\cite{wu2026mefdetr}, instead obtain context through multi-scale feature fusion. RF-DETR interpolates patch embeddings as FlexiViT does~\cite{beyer2023flexivit}, notes that its segmentation features are tied to the input resolution, and observes that a larger patch at a larger input decouples the number of tokens from the segmentation-feature resolution~\cite[App.~L]{robinson2025nasdetr}. The proposed DGC detector decouples them differently: a parameter-free resampling front keeps the pretrained patch embedding, and single-window attention computes global context on the compact grid while masks are supervised and decoded in the larger frame.

Real-time instance segmentation has developed along one-stage designs (Supplementary Section~S1) and along detection transformers with mask heads such as Mask DINO~\cite{li2023maskdino} and RF-DETR-Seg~\cite{robinson2025nasdetr}, whose mask head follows Mask DINO.

\subsection{Mask refinement and mask quality}

Inside the detector, boundary quality has been improved by boundary-preserving branches~\cite{cheng2020bmask}, point-wise rendering~\cite{kirillov2020pointrend}, and transformers that refine incoherent regions~\cite{ke2022masktransfiner}. After the detector, model-agnostic refiners correct its masks. SegFix~\cite{yuan2020segfix} replaces the labels of boundary pixels with those of nearby interior pixels, SegRefiner~\cite{wang2023segrefiner} refines masks with a discrete diffusion process, SAMRefiner~\cite{lin2025samrefiner} turns coarse masks into prompts for a foundation segmenter, and boundary patch refinement (BPR)~\cite{tang2021bpr} predicts a new binary mask for patches cropped along predicted boundaries. Of these, BPR is the closest to the proposed PBR: both refine patches along predicted boundaries with a network fitted on a fixed detector's own training predictions. PBR differs in what it reads and predicts: it reads the detector's soft posterior instead of a binary mask, adds a correction to the mask logits through an output layer initialised to zero, so that it starts as the identity, and has 47.7\,k parameters, far fewer than BPR's high-resolution segmentation networks; Section~\ref{sec:abl-pbr} tests which of these properties matter. Mask Scoring R-CNN~\cite{huang2019maskscoring} showed that ranking masks by an estimate of their quality, in addition to class confidence, raises AP, and SOLO and Mask2Former multiply the class score by the mean posterior inside the mask~\cite{wang2022solo,cheng2022mask2former}. Our quality score Q keeps this training-free form but replaces the mean by Nowozin's approximation of the expected IoU~\cite{nowozin2014iou}, which also penalises posterior mass outside the mask, and computes it on the corrected posterior, so that boundary correction and ranking act on the same estimate.

\subsection{Classification refinement and inference-time components}

Decoupled classification refinement (DCR)~\cite{cheng2018dcr} showed that a separately trained classifier applied to detector outputs can correct errors that arise when localisation and classification share one feature. Open-vocabulary detectors classify region crops with a frozen encoder and fuse the result with the detector's scores by a weighted geometric mean~\cite{gu2022vild,kuo2023fvlm}. The proposed CPC brings crop classification to closed-set, fine-grained categories with a linear probe on frozen DINOv2 features, keeps every mask fixed, and reads at most three crops per image. Its main difference is how the probe's evidence enters the scores. DCR takes the plain product of the two scores, and relabelling replaces the mask's category; CPC instead keeps the detector's hypotheses and chooses between soft fusion and a confident-disagreement gate with a rule evaluated on validation data (Section~\ref{sec:abl-cpc} compares these rules on the same crops). Our post-training components draw on three ideas: averaging the weights of fine-tuned models improves accuracy at no inference cost, as shown for image and text classifiers~\cite{wortsman2022modelsoups}; DDQ trains detection transformers with class-agnostic non-maximum suppression among their queries~\cite{zhang2023ddq}; and SFA averages the predictions of all decoder layers into a reference output for a consistency loss during training~\cite{wang2021sfa}. W, P, and L bring these ideas to a detection transformer at inference, without retraining (Section~\ref{sec:method-recipe}).
\section{Method}
\label{sec:method}

\begin{figure*}[pos=p]
\centering
\includegraphics[width=0.96\textwidth]{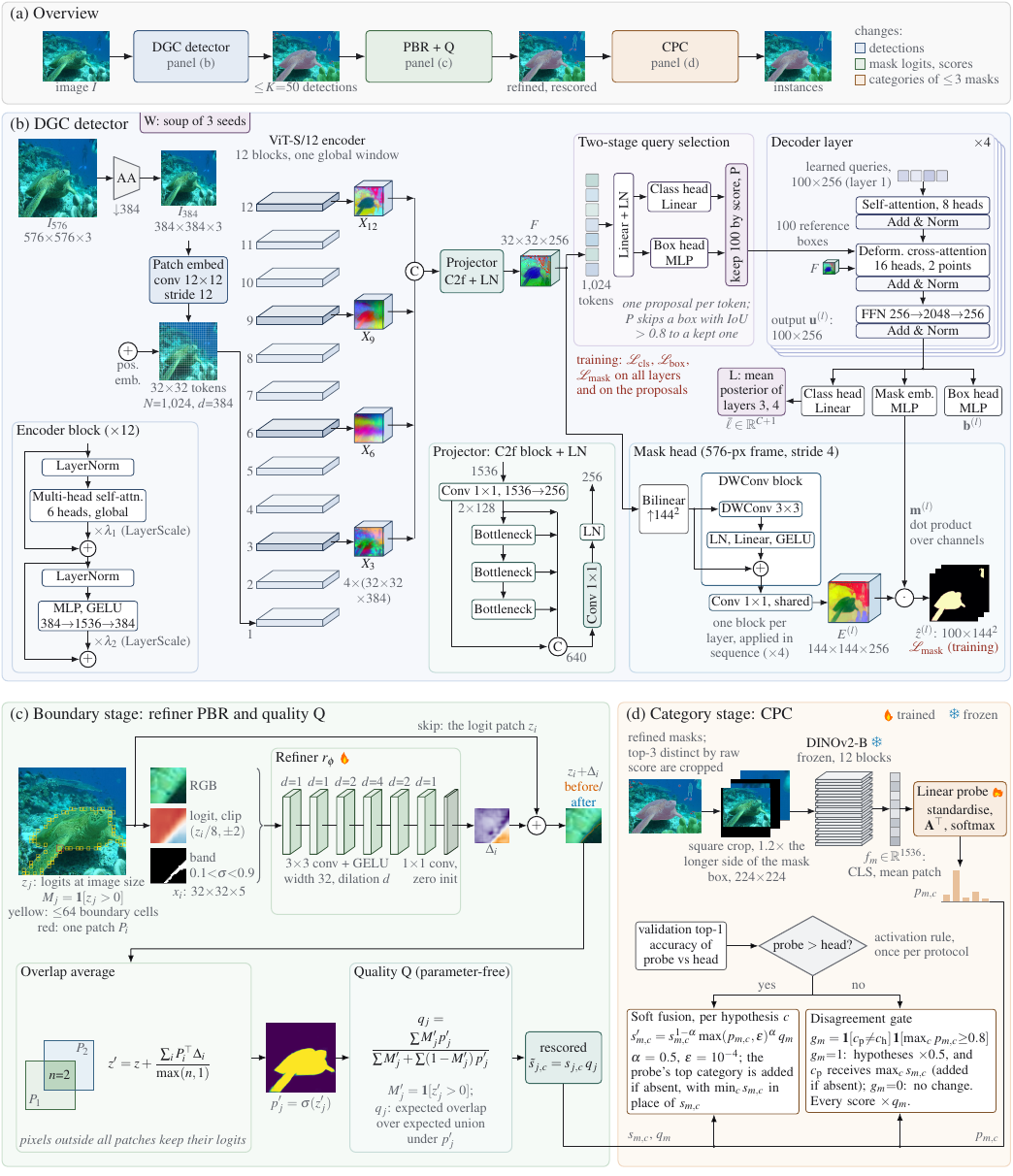}%
\caption{Framework of \method{}: Decoupled Global Context and Selective Boundary-Category Refinement for Underwater Instance Segmentation. (a) Overview: the three stages and the output each changes. (b) The DGC detector: the 576-pixel image is resampled to 384 pixels (AA: antialiased), all twelve encoder blocks attend globally, a C2f projector fuses four block outputs into the memory $F$, 100 selected proposals give the reference boxes of learned content queries, and the decoder and mask head work in the 576-pixel frame (Section~\ref{sec:method-dgc}). Purple: our post-training components W (weight soup of the three seed models), P (proposal selection), and L (layer averaging); C: concatenation; $\odot$: dot product over channels; red: training-only losses. (c) The boundary stage: the refiner $r_\phi$ of PBR corrects the mask logits in $32\times32$ patches $P_i$ on mask boundaries, averaged where $n$ patches overlap, and Q rescores every detection. (d) The category stage: at most three distinct masks per image, those with the highest raw score, are encoded by a frozen DINOv2-B and classified by a linear probe, and a rule fixed once per protocol on validation data selects soft fusion or the disagreement gate. Feature maps (first three principal components), image patches, and maps are real tensors; the overlap sketch, feature vector, and bar chart are schematic.}
\label{fig:framework}
\end{figure*}

\subsection{Overview}
\label{sec:method-overview}

Given an image $I$, \method{} returns a set of instances, each with a binary mask, a category, and a score. It proceeds in three stages that change different outputs (Fig.~\ref{fig:framework}a). Fig.~\ref{fig:framework}b details the detector, and Fig.~\ref{fig:framework}c and~d the two refinement stages. The detector produces at most $K=50$ detections, query--category pairs $(j,c)$ with a score $s_{j,c}$; the pairs of query $j$ share its mask-logit field $\hat z_j$ and box $\mathbf b_j$ and are the category hypotheses of that mask (Section~\ref{sec:method-dgc}). The boundary stage replaces each mask-logit field by a corrected one and rescores the detections with a quality estimate. The category stage revises the category scores of at most three masks per image and changes no mask. The detector and the boundary refiner are trained on the training split of each protocol, the category probe on ground-truth crops of the same split, and the activation rule is evaluated once on validation data. Supplementary Table~S5 lists, for each proposed component, the related work and what is new in it.

\subsection{Decoupled global context (DGC) detector}
\label{sec:method-dgc}

\paragraph{Base detector} The base detector is the segmentation variant of RF-DETR~\cite{robinson2025nasdetr} at its Small operating point (Fig.~\ref{fig:framework}b). A ViT-S encoder initialised from DINOv2~\cite{oquab2023dinov2} embeds non-overlapping $12\times12$ patches; at 384 pixels it processes a $32\times32$ grid of 384-dimensional tokens. Nine of its twelve blocks attend within $2\times2$ windows, and blocks 4, 7, and 10 attend globally. The outputs of blocks 3, 6, 9, and 12 are concatenated and fused by the C2f projector of LW-DETR~\cite{chen2024lwdetr} into a single-scale memory $F$ of $32\times32\times256$. Two-stage query selection~\cite{zhu2021deformabledetr}, in the mixed form of DINO~\cite{zhang2023dino}, scores one proposal per memory token and keeps the $N_{\mathrm q}=100$ highest as the reference boxes of learned content queries, and four decoder layers with self-attention, deformable cross-attention to $F$, and a feed-forward network refine the queries. After every layer $l$, shared heads predict category logits $\ell^{(l)}_j\in\mathbb R^{C+1}$ (the $C$ categories and a no-object class), a box, and a mask embedding; as in Mask DINO~\cite{li2023maskdino}, the mask logits are the dot product of the mask embedding with a pixel embedding, obtained by upsampling $F$ to stride~4 of the input frame and refining it with one depthwise-convolution block per layer. At inference, with $\bar\ell_j$ the category logits of query $j$ after layer averaging (Section~\ref{sec:method-recipe}) and $\sigma(x)=1/(1+\exp(-x))$, the scores $s_{j,c}=\sigma(\bar\ell_{j,c})$ of all query--category pairs are ranked, the $K=50$ highest with $s_{j,c}>10^{-3}$ are kept, and pairs of the no-object class are dropped. The image is resized to the square input frame without preserving its aspect ratio, and the mask logits of the kept queries are resampled bilinearly back to the image; we write $z_j$ for this native-resolution field and $M_j=\ind[z_j>0]$ for the mask.

\begin{figure*}[pos=t]
\centering
\includegraphics[width=\textwidth]{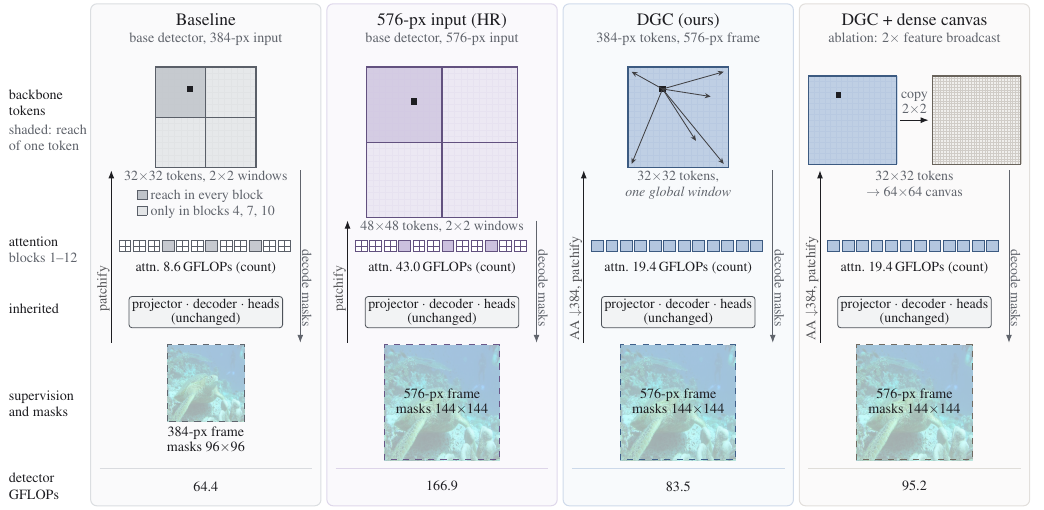}
\caption{What the DGC detector decouples. Top: the backbone's token grid, with the reach of one token in every block (dark) and only in the global blocks 4, 7, and 10 (light), and the attention of its twelve blocks (filled: global; quartered: $2\times2$ windows), with the attention GFLOPs of each network (encoder blocks and decoder self-attention) from a direct count. Bottom: the frame in which the detector is supervised and decodes masks, and the GFLOPs of the whole detector. Accuracy and engine time: Section~\ref{sec:abl-dgc}.}
\label{fig:dgc}
\end{figure*}

\paragraph{Decoupling} Raising the input from 384 to 576 pixels gives this detector its largest gain (Section~\ref{sec:abl-dgc}), but the backbone then processes $48\times48=2{,}304$ tokens instead of 1{,}024. The base design ties two quantities to the input resolution: the token grid on which the backbone computes context, and the frame in which the model is supervised and decodes masks. RF-DETR observes that its segmentation features are tied to the input resolution and that a larger patch at a larger input unties the number of tokens from the segmentation-feature resolution~\cite[App.~L]{robinson2025nasdetr}; at 576 pixels this would need an 18-pixel patch and a new patch embedding. The proposed DGC detector instead decouples the two with a parameter-free resampling front that keeps the pretrained 12-pixel patch embedding, and pairs it with single-window attention (the setting of RF-DETR-Seg Nano), so that every block attends globally:
\begin{equation}
\begin{aligned}
I_{384}&=\mathcal{R}^{\mathrm{AA}}_{384}(I_{576}),\\
F&=B_{\mathrm{glob}}(I_{384}),\\
\big(\hat z^{(l)},\ell^{(l)},\mathbf b^{(l)}\big)_{l=1}^{4}&=H_{576}(F).
\end{aligned}
\label{eq:dgc}
\end{equation}
Here, $I_{576}$ is the image resized to $576\times576$ and $\mathcal R^{\mathrm{AA}}_{384}$ is a parameter-free antialiased bilinear resampling to 384 pixels. $B_{\mathrm{glob}}$ is the backbone with a single attention window followed by the projector: each of its twelve blocks applies six-head self-attention (head width 64) over all 1{,}024 patch tokens and the class token, and $F$ is the $32\times32\times256$ memory. $H_{576}$ is the inherited query selection, decoder, and heads, working in the 576-pixel frame: training images and their masks are stored with the longer side capped at 576 pixels (384 for a 384-pixel detector) and resized to $576\times576$, the pixel embedding and the mask logits are $144\times144$ instead of $96\times96$, and the mask losses compare point samples of these logits with 576-pixel ground truth; normalised box targets are the same in both frames. Fig.~\ref{fig:dgc} contrasts four of the configurations studied in Section~\ref{sec:abl-dgc}.

\paragraph{Cost of attention} Global attention is affordable on the compact grid. With $N$ tokens of width $d$ split into $W_l$ windows in block $l$, the two matrix products of attention cost
\begin{equation}
C_{\mathrm{attn}}=4\,d\sum_{l=1}^{12}\frac{N^2}{W_l}\ \text{FLOPs}.
\label{eq:attn-cost}
\end{equation}
For $d=384$, this gives 8.5\,GFLOPs for the baseline ($N=1{,}024$, nine blocks with $W_l=4$), 19.3 for the DGC backbone ($N=1{,}024$, $W_l=1$), and 42.8 for the 576-pixel detector ($N=2{,}304$, nine blocks with $W_l=4$); a direct count of each network, which also includes the class token of every window and the decoder's self-attention, gives 8.6, 19.4, and 43.0. GFLOPs in this paper count a multiply--add as two operations; deformable sampling, resampling, softmax, and normalisation are not counted. The token-wise layers, which dominate the backbone, scale with $N$ and stay at the baseline's cost. The whole DGC detector costs 83.5\,GFLOPs, half of the 576-pixel detector's 166.9, and it keeps the base detector's parameters (33.42\,M for ten categories, of which 2.7\,M are training-only query groups).

\paragraph{Training} Each detector is fine-tuned from the COCO-pretrained RF-DETR-Seg Small checkpoint on the training split for 20 epochs, once with each of three random seeds (17, 29, 43); we call the three resulting networks seed models. Training uses an adapted RF-DETR code base on two T4 GPUs in mixed precision (effective batch 16). AdamW~\cite{loshchilov2019adamw} (weight decay $10^{-4}$, none on biases, normalisation layers, and embeddings) uses peak learning rates of $10^{-4}$ for the projector, query selection, and heads, $0.7\times10^{-4}$ for the decoder, and $1.5\times10^{-4}\cdot0.7^{2}\cdot0.8^{\,14-i}$ for encoder block $i$ (layer-wise decay 0.8: $4.0\times10^{-6}$ for block~1 to $4.7\times10^{-5}$ for block~12), with gradient-norm clipping at 0.1, a two-epoch linear warm-up, and cosine decay to 5\,\% of the peak. Thirteen query groups are trained, as in Group DETR~\cite{chen2023groupdetr}, and inference keeps the first. The inherited losses and an exponential moving average of the weights (decay 0.993) are kept; validation mask AP is measured every two epochs (every epoch on USIS16K), and the better of the live and the averaged weights at its best epoch is retained for each seed model. Supplementary Table~S3 lists the augmentation and the remaining settings. Every detector in this paper, including the ablation arms, uses this training configuration.

\subsection{Weight soup, proposal selection, and layer averaging}
\label{sec:method-recipe}

Three inexpensive post-training components of \method{} complete the detector at inference. They are part of every Full model in this paper, including the Full configurations of the base and 576-pixel detectors. We call a detector without them native; the Baseline results exclude them as well.
\paragraph{Weight soup (W)} W averages the checkpoints $\theta_1$, $\theta_2$, and $\theta_3$ of the three seed models in double precision, as model soups do for fine-tuned classifiers~\cite{wortsman2022modelsoups},
\begin{equation}
\theta_{\mathrm W}=\frac{1}{3}\sum_{k=1}^{3}\theta_k,
\label{eq:soup}
\end{equation}
which yields one network at the cost of one detector. The three seed models are trained for the evaluation in any case, so W adds no training, whereas deploying a single seed model would give up the soup's gain. Averaging needs corresponding parameters: the learned queries of a detection transformer form an unordered set, so independently trained models could give one role to different query slots. The seed models are fine-tuned from one COCO checkpoint and keep their slots aligned; Section~\ref{sec:abl-stack} measures both the gain and the alignment.
\paragraph{Proposal selection (P)} P selects distinct proposals at inference only, by class-agnostic non-maximum suppression among the encoder proposals at an IoU of 0.8, the selection with which DDQ trains its detectors~\cite{zhang2023ddq}. Our detector is trained with the usual top-$k$ selection, so P needs no retraining. Let $\pi_1,\pi_2,\dots$ order the encoder proposals by score, with boxes $\mathbf b_{\pi_k}$. Starting from $\mathcal S_0=\emptyset$, a proposal joins the selected set unless it overlaps one already selected:
\begin{equation}
\mathcal S_k=
\begin{cases}
\mathcal S_{k-1}\cup\{\pi_k\}, & \displaystyle\max_{j\in\mathcal S_{k-1}}\operatorname{IoU}\big(\mathbf b_{\pi_k},\mathbf b_j\big)\le\tau_{\mathrm P},\\[2pt]
\mathcal S_{k-1}, & \text{otherwise},
\end{cases}
\label{eq:proposal}
\end{equation}
with $\tau_{\mathrm P}=0.8$ and the maximum over the empty set taken as zero. Selection stops when $N_{\mathrm q}$ proposals are selected; if fewer pass, the highest-scoring unselected proposals complete the set.
\paragraph{Layer averaging (L)} L averages the category posteriors of the last two decoder layers at inference. SFA averages all decoder layers into a reference output during training~\cite{wang2021sfa}; L needs no training. With $\ell^{(l)}_j$ the category logits of query $j$ after decoder layer $l$, the final logits are those of the mean posterior of the last two layers,
\begin{equation}
\bar\ell_j=\operatorname{logit}\Big(\frac{1}{2}\sum_{l=3}^{4}\sigma\big(\ell^{(l)}_j\big)\Big),
\label{eq:layeravg}
\end{equation}
with the mean posterior clamped to $[10^{-6},1-10^{-6}]$; masks and boxes are unchanged.

\subsection{Posterior-conditioned boundary refinement (PBR)}
\label{sec:method-pbr}

For each mask of a kept detection with score $s_{j,c}\ge0.02$, PBR marks the $16\times16$-pixel cells that contain a boundary pixel of $M_j$, that is, a pixel removed by a $3\times3$ erosion (Fig.~\ref{fig:framework}c); masks below the threshold keep $z'_j=z_j$. It keeps at most 64 cells per mask, subsampled evenly, and centres a $32\times32$ patch on each, shifted inside the image at the borders; $P_i$ crops patch $i$ from an image-sized field and $P_i^{\top}$ places a patch back. The refiner $r_\phi$ receives a five-channel input $x_i$ and predicts a logit correction $\Delta_i$:
\begin{equation}
\begin{aligned}
x_i&=\Big[\,I_i,\;\operatorname{clip}\!\big(z_i/8,\,-2,\,2\big),\;\ind\big[\,0.1<\sigma(z_i)<0.9\,\big]\Big],\\
\Delta_i&=r_\phi(x_i),
\end{aligned}
\label{eq:pbr}
\end{equation}
where $I_i=P_iI$ is the normalised RGB patch (three channels) and $z_i=P_iz_j$; the fourth channel is the detector's mask logit, scaled and clipped (a monotone transform of its posterior), and the fifth marks the uncertainty band of $\sigma(z_i)$, in the spirit of PointRend's selection of uncertain points~\cite{kirillov2020pointrend}. The network $r_\phi$ has six $3\times3$ convolution-GELU layers of width 32 with dilations $(1,1,2,4,2,1)$ and a $1\times1$ output layer initialised to zero, so that $z_i+r_\phi(x_i)$ starts as the identity; it has 47{,}745 parameters and costs 0.097\,GFLOPs per patch. Overlapping corrections are averaged, with $n=\sum_iP_i^{\top}\mathbf 1$ the number of patches covering each pixel $u$:
\begin{equation}
z'(u)=z(u)+\frac{\sum_i\big[P_i^{\top}\Delta_i\big](u)}{\max\big(n(u),1\big)},\qquad M'=\ind[z'>0].
\label{eq:pbr-avg}
\end{equation}
Pixels outside all patches keep their logits, and inside a patch the correction is free to act beyond the uncertainty band, which enters as an input.

The refiner is fitted per protocol on the fixed detector's own predictions for 3{,}000 training images. Each non-crowd ground-truth instance is matched to the query of the detector without P and L whose box has the highest IoU with it, when that IoU is at least 0.5; twelve patches centred on randomly drawn boundary pixels are taken per matched instance, and the target is the ground-truth mask on the patch. Training minimises binary cross-entropy on $z_i+\Delta_i$ with AdamW (learning rate $10^{-3}$, weight decay $10^{-4}$, batch 512, 12 epochs) and takes minutes on one GPU. Like BPR~\cite{tang2021bpr}, which is also fitted on a fixed detector's training predictions and transfers to other detectors, PBR refines boundary patches after the detector; BPR predicts a new binary mask from the RGB patch and the binary mask with a high-resolution network, whereas PBR conditions on the soft posterior and an uncertainty band and predicts a residual on the logits with 47.7\,k parameters.

\subsection{Quality from the corrected mask}
\label{sec:method-q}

The category score of a mask detector ranks hypotheses by class confidence, which can disagree with mask quality~\cite{huang2019maskscoring}. SOLO and Mask2Former multiply the class score by the mean posterior inside the mask~\cite{wang2022solo,cheng2022mask2former}. \method{} multiplies each score by the ratio of the expected intersection to the expected union under independent pixel posteriors, Nowozin's approximation of the expected IoU~\cite{nowozin2014iou}, computed on the corrected posterior $p'_j=\sigma(z'_j)$ of each mask:
\begin{equation}
q_j=\frac{\sum_u M'_j(u)\,p'_j(u)}{\sum_u M'_j(u)+\sum_u\big(1-M'_j(u)\big)\,p'_j(u)},\quad \tilde s_{j,c}=s_{j,c}\,q_j,
\label{eq:q}
\end{equation}
with the denominator floored at $10^{-6}$. Without the second term of its denominator, $q_j$ is the mean posterior inside the mask; the second term penalises posterior mass outside the output mask, which makes $q_j$ the soft Jaccard index between $p'_j$ and $M'_j$. If the pixels of the true instance are drawn independently from $p'_j$, the numerator is the expected overlap between $M'_j$ and the instance and the denominator the expected union; their ratio approximates the expected IoU, with an error that decays with the number of pixels~\cite{nowozin2014iou}, and carries no calibration guarantee. Computing $q_j$ on $z'_j$ keeps the ranking consistent with the geometry the model outputs. The complete configuration W + P + L + PBR + Q is called \emph{Full}.

\subsection{Crop-probe category refinement (CPC)}
\label{sec:method-cpc}

\paragraph{Selection and features} In each image the distinct masks of the kept detections are ranked by their highest raw category score, and the top three (or all of them, if there are fewer) are selected (Fig.~\ref{fig:framework}d); every other detection keeps its Full score. For a selected mask $m$, a square crop of $1.2\times$ the longer side of the box of the refined mask is taken around the box centre, with zero padding outside the image, resized to $224\times224$ (bicubic), normalised with the ImageNet statistics, and encoded by a frozen DINOv2-B~\cite{oquab2023dinov2}. The feature $f_m\in\mathbb R^{1536}$ concatenates the class token and the mean patch token, as in DINOv2's linear evaluation~\cite{oquab2023dinov2}.

\paragraph{Probe} A linear softmax probe maps standardised features to category posteriors:
\begin{equation}
p_m=\textup{softmax}\!\big(\mathbf A^{\top}\big((f_m-\mu)\oslash\sigma_f\big)+\mathbf a_0\big).
\label{eq:probe}
\end{equation}
Here, $\mu$ and $\sigma_f$ are the mean and the standard deviation (plus $10^{-6}$) of the training features, $\oslash$ is element-wise division, and $\mathbf A\in\mathbb R^{1536\times C}$ and $\mathbf a_0\in\mathbb R^{C}$ are fitted on ground-truth crops of the training split by $\ell_2$-penalised multinomial logistic regression (L-BFGS), with the penalty chosen from $\{10^{-5},\dots,10^{-2}\}$ by the held-out log-loss of five-fold, image-grouped cross-validation. The probe is independent of the detector.

\paragraph{Soft fusion} For every category hypothesis $c$ of a selected mask $m$, with raw detector score $s_{m,c}$ and quality $q_m$, soft fusion takes the weighted geometric mean of the detector and crop-classifier scores used by open-vocabulary detectors~\cite{gu2022vild,kuo2023fvlm}:
\begin{equation}
s'_{m,c}=s_{m,c}^{\,1-\alpha}\,\max\!\big(p_{m,c},\,\epsilon\big)^{\alpha}\,q_m.
\label{eq:soft}
\end{equation}
We set $\alpha=0.5$ and $\epsilon=10^{-4}$.
If the probe's top category is absent from the mask's hypotheses, it is added with the mask's lowest detector score in place of $s_{m,c}$.

\paragraph{Disagreement gate} Let $c_{\mathrm h}$ be the mask's highest-scoring detector category, $c_{\mathrm p}=\arg\max_c p_{m,c}$ the probe's, and $o_m=\max_c s_{m,c}$. The gate
\begin{equation}
g_m=\ind\big[c_{\mathrm p}\neq c_{\mathrm h}\big]\;\ind\big[\max_c p_{m,c}\ge\tau_{\mathrm g}\big],
\label{eq:gate}
\end{equation}
with $\tau_{\mathrm g}=0.8$, leaves the Full scores $s_{m,c}q_m$ unchanged when $g_m=0$. When $g_m=1$, the scores become
\begin{equation}
s'_{m,c}=q_m\cdot\begin{cases}o_m, & c=c_{\mathrm p},\\ \lambda_{\mathrm g}\,s_{m,c}, & c\neq c_{\mathrm p},\end{cases}
\label{eq:gate-update}
\end{equation}
with $\lambda_{\mathrm g}=0.5$: the detector's hypotheses are kept rather than discarded, and the probe's category is added if it is absent. Objects with $g=0$ keep their label, so for a hard switch from $c_{\mathrm h}$ to $c_{\mathrm p}$ on objects of true category $y$, the change in top-1 accuracy is exactly
\begin{equation}
\begin{aligned}
\Delta\mathrm{Acc}=\Pr(g=1)\,\Big[&\Pr\big(c_{\mathrm p}=y\mid g=1\big)\\
&-\Pr\big(c_{\mathrm h}=y\mid g=1\big)\Big].
\end{aligned}
\label{eq:gate-identity}
\end{equation}
Hence, the switch raises top-1 accuracy whenever the probe is more often right than the head on the objects where the two confidently disagree, even if the probe is the weaker classifier on average.

\paragraph{Activation rule} Soft fusion is used on a protocol if and only if the probe's top-1 accuracy exceeds the detector head's on the validation objects $\mathcal O_{\mathrm{val}}$,
\begin{equation}
\sum_{o\in\mathcal O_{\mathrm{val}}}\Big(\ind\big[c_{\mathrm p}(o)=y_o\big]-\ind\big[c_{\mathrm h}(o)=y_o\big]\Big)>0;
\label{eq:rule}
\end{equation}
otherwise the gate is used. Each non-crowd ground-truth object $o$ of category $y_o$ is matched to the highest-scoring distinct mask with mask IoU of at least 0.5, and the head's category is that mask's top-scoring one. The rule compares two classifiers on the same objects and involves no AP measurement; it was evaluated with a flip-averaged variant of the probe (two views per crop), whereas the deployed probe reads one view. The class-agnostic USIS10K task has a single category and no category branch. UIIS has no validation partition separate from its test partition, so it receives the default, the gate.

\subsection{Complexity}
\label{sec:method-cost}

The inference cost of the complete model decomposes as
\begin{equation}
C_{\mathrm{tot}}=C_{\mathrm{det}}+N_{\mathrm{patch}}\,C_{\mathrm{PBR}}+\min(3,N_{\mathrm{mask}})\,C_{\mathrm{crop}},
\label{eq:cost}
\end{equation}
where $N_{\mathrm{mask}}$ is the number of distinct masks of the kept detections and $N_{\mathrm{patch}}$ the number of their boundary patches (detections that share a mask are refined once), $C_{\mathrm{det}}=83.5$\,GFLOPs for the DGC detector, $C_{\mathrm{PBR}}=0.097$\,GFLOPs per boundary patch, and $C_{\mathrm{crop}}=46.3$\,GFLOPs per DINOv2-B crop (12.3 for DINOv2-S); W adds no cost, and P, L, and Q are negligible. With 133 to 1{,}072 boundary patches per image, the complete model costs 235 to 327\,GFLOPs per image on the test images used for timing, against 420 to 2{,}775\,GFLOPs for the networks of the competing methods (Section~\ref{sec:exp-eff}).
\section{Experiments}
\label{sec:experiments}

\subsection{Datasets and protocols}
\label{sec:exp-data}

Experiments cover four image corpora and five protocols (Table~\ref{tab:datasets}). UIIS~\cite{lian2023watermask} has seven categories, and its 691-image evaluation partition serves as both validation and test in the published comparisons. UIIS10K~\cite{li2025uwsam} has ten categories and a 2{,}010-image test split; it provides no validation split, so a fixed 10\,\% of its training images (804) serves for validation. USIS10K~\cite{lian2024usissam} is evaluated in both of its label spaces, the class-agnostic salient-instance task (USIS10K-A) and seven categories (USIS10K-M), which share their images. USIS16K~\cite{hong2025usis16k} has 158 fine-grained categories; the released split has 11{,}242 training, 1{,}539 validation, and 3{,}370 test images. Its dataset paper describes a 1{,}615-image test partition, not the released test folder, and leaves the task variant of its benchmark ambiguous, so the released test folder is scored with categories, the stricter variant.

\subsection{Metrics and statistical analysis}
\label{sec:exp-metrics}

The main metric is COCO mask AP~\cite{lin2014coco} averaged over IoU thresholds 0.50:0.95, computed with pycocotools at the original image resolution, together with AP$_{50}$, AP$_{75}$, and AP by object size. Object areas are mask areas for the ground truth and for the detections (results are scored without their boxes, as in Detectron2 and MMDetection). Size strata use COCO areas, except on USIS16K, whose benchmark defines them by the fraction of the image area ($<5\,\%$, $5$--$30\,\%$, $>30\,\%$). One evaluator scores every model in this paper, including the competing methods, and re-scoring the stored predictions reproduces every reported AP exactly. Supplementary Table~S2 adds mask AR, Boundary AP~\cite{cheng2021boundaryiou}, and IoU-based measures for each model scored with this evaluator. Its IoU measures depend on a score threshold. Because Q multiplies every score by a quality below one, fewer of our masks pass a fixed threshold: at a threshold of 0.5, \method{} trails DiveSeg in foreground IoU on UIIS, UIIS10K, and USIS10K-M and in mIoU on UIIS and UIIS10K, whereas its masks that match an annotation are more accurate than those of every competing method on all five protocols.

Differences between models are estimated with a paired image bootstrap~\cite{efron1993bootstrap}. Each of $B=600$ draws resamples the test images with replacement (seed 17), recomputes the AP of both models on the same resample, and records the difference; tables report the point difference and the percentile 95\,\% interval, and \sig{} marks an interval that excludes zero. Differences are computed before rounding. Pooled contrasts over protocols average the per-protocol draws; because USIS10K-A and USIS10K-M share their images, a pooled interval that includes both is approximate. The intervals are not corrected for multiplicity, so a bound close to zero is weak evidence, and they describe variation over test images for fixed models; seed-to-seed variation is reported separately through the three native seed models.

Every hyperparameter of the detector, the refiner, and the category branch was selected on validation data, and during development the training code was blocked from reading the official test partitions of UIIS10K, USIS10K, and USIS16K. UIIS has no validation data apart from its 691-image evaluation partition: as in the published UIIS comparisons, each seed model's checkpoint was selected on that partition, while the refiner, the probe, the category rule, and all other settings were transferred unchanged from the other protocols. The DGC configuration was adopted after a test comparison with its dense-canvas variant (Section~\ref{sec:abl-dgc}), and it also ranks above that variant on validation. Most test contrasts were fixed in advance, together with their models and decision rules; the canvas contrast within the 576-pixel frame and the comparison of the DGC detector with the baseline in Section~\ref{sec:abl-dgc} were added afterwards.

\begin{table*}[width=\textwidth,cols=9,pos=t]
\caption{Mask AP (\%, IoU 0.50:0.95) on the official test partitions of the five protocols.}\label{tab:sota}
\scriptsize
\renewcommand{\arraystretch}{0.8}
\setlength{\tabcolsep}{2.6pt}
\begin{tabular*}{\tblwidth}{@{\extracolsep{\fill}}lllrrrrrr@{}}
\toprule
Method & Venue & Backbone & Params (M)$\downarrow$ & UIIS$\uparrow$ & UIIS10K$\uparrow$ & USIS10K-A$\uparrow$ & USIS10K-M$\uparrow$ & USIS16K$\uparrow$ \\
\midrule
\multicolumn{9}{@{}l}{\textit{Generic instance and salient-instance segmentation (as benchmarked on the underwater datasets; strongest per protocol)}} \\
\quad Mask Scoring R-CNN~\cite{huang2019maskscoring,li2025uwsam,lian2023watermask} & CVPR '19 & R101 & 75.68 & 24.6 & 37.3$^{\dagger}$ & -- & -- & -- \\
\quad PointRend~\cite{kirillov2020pointrend,li2025uwsam}$^{\dagger}$ & CVPR '20 & R50 & 53.66 & -- & 37.3 & -- & -- & -- \\
\quad PointRend~\cite{kirillov2020pointrend,hong2025usis16k}$^{\dagger}$ & CVPR '20 & -- & 64.69 & -- & -- & -- & -- & 76.3 \\
\quad QueryInst~\cite{fang2021queryinst,li2025uwsam,lian2023watermask} & ICCV '21 & R101 & 182.6 & 26.0 & 33.9$^{\dagger}$ & -- & -- & -- \\
\quad RDPNet~\cite{wu2021rdpnet,lian2024usissam} & TIP '21 & R101 & -- & -- & -- & 54.7 & 39.3 & -- \\
\quad OQTR~\cite{pei2023oqtr,lian2024usissam} & TMM '23 & R50 & -- & -- & -- & 56.6 & 19.7 & -- \\
\midrule
\multicolumn{9}{@{}l}{\textit{Foundation-model (SAM) based}} \\
\quad Mask R-CNN + SAM~\cite{he2017maskrcnn,kirillov2023sam,lian2024usissam} & ICML '24 & ViT-H & -- & -- & -- & 55.1 & 38.5 & -- \\
\quad USIS-SAM~\cite{lian2024usissam,hong2025usis16k} & ICML '24 & ViT-H & 698.9 & 29.4 & -- & 59.7 & 43.1 & 81.0$^{\dagger}$ \\
\quad USIS-SAM~\cite{lian2024usissam,li2025uwsam}$^{\dagger}$ & ICML '24 & ViT-H (adapter) & 665.8 & -- & 39.8 & -- & -- & -- \\
\quad UWSAM (student)~\cite{li2025uwsam}$^{\dagger}$ & arXiv '25 & ViT-S & 47.39 & 26.3 & 38.7 & -- & 41.0 & -- \\
\quad UWSAM (teacher)~\cite{li2025uwsam}$^{\dagger}$ & arXiv '25 & ViT-H (LoRA) & 632.3 & 31.9 & 44.6 & -- & 46.0 & -- \\
\quad BiPA~\cite{ma2026bipa} & CVPR '26 & ViT-H + R50 & 737 & 32.1 & -- & 64.2 & 45.2 & -- \\
\midrule
\multicolumn{9}{@{}l}{\textit{Underwater-specific}} \\
\quad WaterMask R-CNN (1$\times$)~\cite{lian2023watermask} & ICCV '23 & R50 & -- & 23.3 & -- & -- & -- & -- \\
\quad WaterMask R-CNN (1$\times$)~\cite{lian2023watermask} & ICCV '23 & R101 & 67 & 25.6 & -- & -- & -- & -- \\
\quad WaterMask R-CNN~\cite{lian2023watermask,lian2024usissam} & ICCV '23 & R50 & -- & 26.4 & -- & 58.3 & 37.7 & -- \\
\quad WaterMask R-CNN~\cite{lian2023watermask,lian2024usissam,li2025uwsam} & ICCV '23 & R101 & 67 & 27.2 & 37.4$^{\dagger}$ & 59.0 & 38.7 & --$^{\ddagger}$ \\
\quad WaterMask R-CNN~\cite{lian2023watermask,hong2025usis16k}$^{\dagger}$ & ICCV '23 & -- & 48.27 & -- & -- & -- & -- & 72.7 \\
\quad Cascade WaterMask R-CNN~\cite{lian2023watermask} & ICCV '23 & R101 & 107 & 27.1 & -- & -- & -- & -- \\
\quad UIS-Mamba~\cite{cong2025uismamba} & ACM MM '25 & UIS-Mamba-T & 56 & 29.4 & -- & 62.2 & -- & -- \\
\quad UIS-Mamba~\cite{cong2025uismamba} & ACM MM '25 & UIS-Mamba-S & 76 & 30.4 & -- & 63.1 & -- & -- \\
\quad UIS-Mamba~\cite{cong2025uismamba} & ACM MM '25 & UIS-Mamba-B & 115 & 31.2 & -- & 63.8 & 46.2 & -- \\
\quad BARD-ERA~\cite{pan2025baris} & ACML '25 & Swin-B & 114.44 & 31.6 & -- & -- & 47.3 & -- \\
\quad BARD-ERA~\cite{pan2025baris} & ACML '25 & ConvNeXt-V2-B & 112.46 & 32.3 & -- & -- & -- & -- \\
\quad Clarity-Former~\cite{chen2026clarityformer} & TGRS '26 & Swin-T & 45 & 27.8 & -- & -- & 42.6 & 80.5 \\
\quad Clarity-Former~\cite{chen2026clarityformer} & TGRS '26 & Swin-S & 67 & 28.9 & --$^{\ddagger}$ & --$^{\ddagger}$ & 45.8 & 82.5 \\
\quad Clarity-Former~\cite{chen2026clarityformer} & TGRS '26 & Swin-B & 106 & 27.2 & -- & -- & 44.3 & 80.3 \\
\quad DiveSeg~\cite{chen2026diveseg} & AAAI '26 & ViT-L & 390 & 35.6 & --$^{\ddagger}$ & 64.1 & 48.4 & --$^{\ddagger}$ \\
\quad ViT-UWA~\cite{jia2026vituwa} & TIP '26 & ViT-UWA-B (IN-22K) & 129 & 29.0 & -- & -- & 44.2 & -- \\
\quad DCFNet~\cite{zheng2026dcfnet} & Symmetry '26 & R50 & 95 & -- & -- & 62.3 & -- & -- \\
\quad DCFNet~\cite{zheng2026dcfnet} & Symmetry '26 & Swin-B & 219 & -- & -- & 63.7 & -- & 85.2 \\
\quad USIS-PGM~\cite{hong2026usispgm} & OCEANS '26 & -- & -- & -- & -- & -- & -- & 81.6 \\
\quad AASNet~\cite{kong2025aasnet}$^{\P}$ & Appl. Sci. '25 & GELAN (YOLOv9) & 27.84 & -- & -- & 65.9 & 47.4 & -- \\
\midrule
\multicolumn{9}{@{}l}{\textit{Reproduced with the authors' code and configurations}} \\
\quad Clarity-Former~\cite{chen2026clarityformer}$^{\ddagger}$ & TGRS '26 & Swin-S & 67 & 28.84 & 40.80 & 61.21 & 41.89 & --$^{\S}$ \\
\quad WaterMask R-CNN~\cite{lian2023watermask}$^{\ddagger}$ & ICCV '23 & R101 & 67 & --$^{\S}$ & 36.07 & 61.00 & 38.98 & 76.98 \\
\quad DiveSeg~\cite{chen2026diveseg}$^{\ddagger}$ & AAAI '26 & ViT-L & 390 & 33.41 & 43.98 & --$^{\S}$ & --$^{\S}$ & 87.95 \\
\midrule
\multicolumn{9}{@{}l}{\textit{Ours}} \\
\quad Baseline$^{a}$ &  & DINOv2-S & 33.4 & 35.13 & 46.04 & 68.04 & 49.47 & 85.72 \\
\quad \method{} &  & DINOv2-S & 33.5 & 37.85 & 49.80 & \underline{70.70} & \underline{52.95} & 87.94 \\
\quad \method{} + CPC &  & DINOv2-S + B & 120.3 & 38.05 & 49.92 & -- & \textbf{53.40} & \underline{90.03} \\
\quad \methodhr{} &  & DINOv2-S & 34.0 & \underline{38.35} & \underline{49.94} & \textbf{71.57} & 52.43 & 88.18 \\
\quad \methodhr{} + CPC &  & DINOv2-S + B & 120.8 & \textbf{38.52} & \textbf{50.01} & -- & 52.82 & \textbf{90.41} \\
\bottomrule
\end{tabular*}
\par\vspace{3pt}\begin{minipage}{\tblwidth}\raggedright\scriptsize Methods discussed in the text; generic methods: the strongest published result per protocol. Supplementary Table~S1 gives every published result, including arXiv preprints other than UWSAM, whose paper introduced UIIS10K. Published values and parameter counts as printed; for a generic method reported by both dataset papers, the parameter count of the UIIS10K paper. Citations: method and reporting papers. Venue: the method's paper (SAM pipelines: the reporting paper). \textbf{Bold} / \underline{underlined}: best / second best. $^{\dagger}$arXiv preprint (on a method name: all of its values). $^{\P}$Partition and evaluator not stated. $^{\ddagger}$Trained by us with the authors' code (one seed). $^{a}$Base detector, three-seed mean, no post-training components. Ours: one model per protocol (three-seed soup); Params include the refiner and, with CPC, the frozen DINOv2-B encoder and the 158-class probe. USIS16K: published rows follow the dataset paper's 1{,}615-image test partition, ours the released 3{,}370-image test folder (Section~\ref{sec:exp-data}). ``--'': not reported; ``--$^{\ddagger}$'': reproduced below; ``--$^{\S}$'': not reproduced (see Table~\ref{tab:same}); no CPC on USIS10K-A (one category).\end{minipage}
\end{table*}
\begin{table}[width=\linewidth,cols=5,pos=t]
\caption{Evaluation protocols (images per split). USIS10K-A and USIS10K-M share their images; UIIS10K's validation split is a fixed 10\,\% of its training images.}\label{tab:datasets}
\scriptsize
\setlength{\tabcolsep}{3pt}
\begin{tabular*}{\tblwidth}{@{\extracolsep{\fill}}lrrrl@{}}
\toprule
Protocol & Train & Val. & Test & Task \\
\midrule
UIIS & 3{,}937 & 691 & 691 & 7 categories; validation = test \\
UIIS10K & 7{,}234 & 804 & 2{,}010 & 10 categories \\
USIS10K-A & 7{,}442 & 1{,}594 & 1{,}596 & salient instances, one class \\
USIS10K-M & 7{,}442 & 1{,}594 & 1{,}596 & same images, 7 categories \\
USIS16K & 11{,}242 & 1{,}539 & 3{,}370 & 158 fine-grained categories \\
\bottomrule
\end{tabular*}
\end{table}

\subsection{Implementation details}
\label{sec:exp-impl}

Training and evaluation used PyTorch and an adapted RF-DETR code base (v1.8.2) on NVIDIA T4 GPUs; Section~\ref{sec:method-dgc} gives the training configuration. Training one seed model takes about 3 to 16 hours on two T4 GPUs, depending on the protocol and the detector, and each soup averages three of them. Three recent underwater methods serve as competing methods, all scored with our evaluator: DiveSeg with a ViT-L backbone~\cite{chen2026diveseg}, Clarity-Former with Swin-S~\cite{chen2026clarityformer}, and WaterMask R-CNN with ResNet-101~\cite{lian2023watermask}. Where the authors released weights for a protocol, those weights are evaluated: DiveSeg on UIIS and both USIS10K protocols, Clarity-Former on USIS16K, and WaterMask R-CNN with the $3\times$ schedule on UIIS. Every other competing model was retrained with the authors' code and configurations and one training seed: Clarity-Former and DiveSeg keep the final model of their schedules (90k and 30k iterations), as their authors report, and WaterMask R-CNN's configuration keeps the best validation checkpoint of its $2\times$ schedule (Supplementary Table~S3). Our detectors keep the best validation checkpoint of each seed model, and each \method{} detector averages its three seed models. Without any of the proposed post-training components, the three native DGC seed models already exceed DiveSeg on four of the five protocols, both on average and individually (Section~\ref{sec:abl-stack}). \methodhr{} applies our post-training components to the base detector reading 576-pixel inputs; the baseline is the base detector at 384 pixels.

\paragraph{Deployment and latency} \method{} runs as a TensorRT 10.16 pipeline on one T4. The detector forms two engines around proposal selection; PBR and the run-length encoding of the final masks run in fused GPU kernels that compute only within each mask's support, Q makes one pass over the image per mask, and CPC crops on the GPU, bit-identical to the reference crops. On every protocol's test split the pipeline with an FP32 detector reproduces the AP of the PyTorch implementation within 0.01 for Full and within 0.04 with CPC, except CPC on UIIS ($+0.31$): there the gate's decisions depend on floating-point differences between GPU kernels, and its gain is $+0.19$ AP with cuDNN's default kernels and $+0.51$ with benchmark-selected kernels. Each competing method runs its own network as a TensorRT engine, built and verified against the release, with the release's pre- and post-processing at a fixed input shape (DiveSeg $480\times640$, Clarity-Former $800\times1088$), which changes their AP by $-3.4$ to $+0.9$ points against Table~\ref{tab:same}; the note of Table~\ref{tab:eff} lists each engine's AP. All methods share one latency protocol: the same test images, a shared JPEG reader, 10 warm-up images, and 200 timed images back to back at batch 1, from upload of the decoded image to final masks, scores, and categories in host memory. Engine latency alone is also measured with a single-artefact benchmark (CUDA events around the engine) and, for our detectors, with \texttt{trtexec}. The runs used separate sessions on T4 GPUs whose clocks could not be locked (median SM clock 810 to 1{,}455\,MHz across runs). The released repository contains this benchmark and its protocol.

\subsection{Comparison with the state of the art}
\label{sec:exp-sota}

Table~\ref{tab:sota} compares \method{} with the mask AP that published methods report in their papers' tables on our five protocols, with each value as printed; for ViT-UWA it shows the configuration that its paper compares with other methods (ImageNet-22K pre-training), and Supplementary Table~S1 lists its other variants. Clarity-Former's published USIS16K result (82.5) comes from its released configuration, which evaluates on the validation folder rather than the test folder; its released weights score 82.59 on the test folder (Table~\ref{tab:same}). Without its category branch, \method{} exceeds the strongest published mask AP on every protocol: by 2.25 on UIIS (DiveSeg, 35.6), 5.20 on UIIS10K (UWSAM teacher, 44.6), 4.80 on USIS10K-A (AASNet, 65.9), 4.55 on USIS10K-M (DiveSeg, 48.4), and 2.74 on USIS16K (DCFNet, 85.2, on the dataset paper's 1{,}615-image partition). With CPC the leads on the category-aware protocols grow to 2.45, 5.32, 5.00, and 4.83. \methodhr{} holds the best value on UIIS (38.52), UIIS10K (50.01), and USIS16K (90.41) with CPC, and on USIS10K-A (71.57), where no category branch applies; \method{} with CPC holds it on USIS10K-M (53.40). The published leaders use 219 to 632\,M parameters, except AASNet (27.8\,M); \method{}'s detector has 33.4\,M, and with the frozen crop encoder the complete model has 120\,M. The base detector alone already exceeds the strongest published result on four protocols (all but UIIS); the proposed detector and post-training components add 2.2 to 3.8 AP over it, and 4.3 with CPC on USIS16K. Supplementary Table~S1 adds AP$_{50}$, AP$_{75}$, and the size strata of every published, competing, and proposed model.

The published results are reported by their authors with their own backbones, pretraining, and training configurations, so a cross-paper gap cannot be attributed to individual stages; Sections~\ref{sec:exp-same} and~\ref{sec:exp-ablation} provide the controlled comparisons. A published UIIS result of 39.0~\cite{wang2026ucissam} addresses camouflaged instance segmentation, a class-agnostic task; under that variant \method{} scores 47.15 on the same test images.

\subsection{Controlled comparison with competing methods}
\label{sec:exp-same}

\begin{table*}[width=\textwidth,cols=9,pos=t]
\caption{Controlled comparison with the competing methods (mask AP, \%, official test).}\label{tab:same}
\scriptsize
\setlength{\tabcolsep}{3pt}
\begin{tabular*}{\tblwidth}{@{\extracolsep{\fill}}llrrrrrrl@{}}
\toprule
Protocol & Method & AP$\uparrow$ & AP$_{50}{\uparrow}$ & AP$_{75}{\uparrow}$ & AP$_S{\uparrow}$ & AP$_M{\uparrow}$ & AP$_L{\uparrow}$ & $\Delta$AP$\uparrow$ vs.\ strongest competitor \\
\midrule
\multirow{9}{*}{UIIS} & WaterMask R-CNN~\cite{lian2023watermask}$^{\S}$ & 26.24 & 43.58 & 28.43 & 9.65 & 21.40 & 37.53 &  \\
 & Clarity-Former~\cite{chen2026clarityformer}$^{\ddagger}$ & 28.84 & 45.92 & 30.89 & 5.33 & 22.95 & 41.45 &  \\
 & DiveSeg~\cite{chen2026diveseg}$^{\S}$ & 35.71 & 52.47 & 38.21 & 6.17 & 26.35 & \underline{53.10} &  \\
 & DiveSeg~\cite{chen2026diveseg}$^{\ddagger}$ & 33.41 & 49.99 & 35.81 & 5.95 & 25.74 & 49.26 &  \\
 & Baseline$^{a}$ & 35.13 & 51.00 & 39.39 & 9.46 & 27.06 & 50.20 &  \\
 & \method{} & 37.85 & 52.66 & 42.20 & 10.87 & 29.40 & 53.05 & $+2.14$ {\scriptsize[$-0.04$, $+4.41$]} \\
 & \quad + CPC & 38.05 & 52.90 & \underline{42.44} & 10.81 & \underline{29.56} & \textbf{53.15} & $+2.33$\sig{} {\scriptsize[$+0.09$, $+4.53$]} \\
 & \methodhr{} & \underline{38.35} & \underline{53.72} & 42.31 & \textbf{12.06} & \textbf{31.16} & 52.32 & $+2.64$ {\scriptsize[$-0.12$, $+5.21$]} \\
 & \quad + CPC & \textbf{38.52} & \textbf{54.00} & \textbf{42.50} & \underline{12.04} & \textbf{31.16} & 52.50 & $+2.80$\sig{} {\scriptsize[$+0.12$, $+5.33$]} \\
\midrule
\multirow{8}{*}{UIIS10K} & WaterMask R-CNN~\cite{lian2023watermask}$^{\ddagger}$ & 36.07 & 50.99 & 40.30 & 8.03 & 27.29 & 45.69 &  \\
 & Clarity-Former~\cite{chen2026clarityformer}$^{\ddagger}$ & 40.80 & 53.76 & 44.27 & 8.05 & 32.69 & 51.23 &  \\
 & DiveSeg~\cite{chen2026diveseg}$^{\ddagger}$ & 43.98 & 56.58 & 47.44 & 8.44 & 34.18 & 56.07 &  \\
 & Baseline$^{a}$ & 46.04 & 58.40 & 50.93 & 13.46 & 37.37 & 56.86 &  \\
 & \method{} & 49.80 & \underline{59.64} & 53.81 & 17.16 & 41.07 & \textbf{59.65} & $+5.82$\sig{} {\scriptsize[$+4.21$, $+7.23$]} \\
 & \quad + CPC & 49.92 & \textbf{59.81} & \textbf{53.96} & 17.67 & 41.59 & \underline{59.64} & $+5.94$\sig{} {\scriptsize[$+4.38$, $+7.31$]} \\
 & \methodhr{} & \underline{49.94} & 58.92 & 53.81 & \underline{18.25} & \textbf{42.27} & 59.30 & $+5.97$\sig{} {\scriptsize[$+4.33$, $+7.53$]} \\
 & \quad + CPC & \textbf{50.01} & 59.01 & \underline{53.90} & \textbf{18.99} & \underline{42.18} & 59.47 & $+6.03$\sig{} {\scriptsize[$+4.45$, $+7.61$]} \\
\midrule
\multirow{6}{*}{USIS10K-A} & WaterMask R-CNN~\cite{lian2023watermask}$^{\ddagger}$ & 61.00 & 82.81 & 70.33 & 21.29 & 53.37 & 67.76 &  \\
 & Clarity-Former~\cite{chen2026clarityformer}$^{\ddagger}$ & 61.21 & 82.83 & 69.15 & 18.28 & 51.98 & 69.66 &  \\
 & DiveSeg~\cite{chen2026diveseg}$^{\S}$ & 65.62 & 83.89 & 74.01 & 18.24 & 53.46 & 75.96 &  \\
 & Baseline$^{a}$ & 68.04 & 88.38 & 78.53 & 28.60 & 58.21 & 76.38 &  \\
 & \method{} & \underline{70.70} & \underline{88.87} & \underline{80.29} & \underline{30.42} & \underline{60.66} & \underline{79.23} & $+5.09$\sig{} {\scriptsize[$+4.22$, $+6.09$]} \\
 & \methodhr{} & \textbf{71.57} & \textbf{89.14} & \textbf{81.84} & \textbf{32.91} & \textbf{62.46} & \textbf{79.32} & $+5.95$\sig{} {\scriptsize[$+5.06$, $+6.89$]} \\
\midrule
\multirow{8}{*}{USIS10K-M} & WaterMask R-CNN~\cite{lian2023watermask}$^{\ddagger}$ & 38.98 & 55.22 & 43.97 & 6.53 & 29.79 & 44.33 &  \\
 & Clarity-Former~\cite{chen2026clarityformer}$^{\ddagger}$ & 41.89 & 56.93 & 46.39 & 7.75 & 30.92 & 47.73 &  \\
 & DiveSeg~\cite{chen2026diveseg}$^{\S}$ & 46.27 & 59.35 & 52.68 & 7.04 & 31.07 & 53.47 &  \\
 & Baseline$^{a}$ & 49.47 & 64.34 & 56.23 & 17.05 & 35.56 & 56.01 &  \\
 & \method{} & \underline{52.95} & \underline{66.34} & \underline{59.99} & \textbf{27.88} & \textbf{38.44} & \underline{59.47} & $+6.69$\sig{} {\scriptsize[$+4.93$, $+8.52$]} \\
 & \quad + CPC & \textbf{53.40} & \textbf{66.82} & \textbf{60.55} & \textbf{27.88} & \underline{38.41} & \textbf{59.96} & $+7.13$\sig{} {\scriptsize[$+5.43$, $+8.83$]} \\
 & \methodhr{} & 52.43 & 65.73 & 58.98 & \underline{20.90} & 37.55 & 58.91 & $+6.16$\sig{} {\scriptsize[$+4.42$, $+7.84$]} \\
 & \quad + CPC & 52.82 & 66.23 & 59.40 & 17.56 & 37.83 & 59.25 & $+6.55$\sig{} {\scriptsize[$+4.74$, $+8.18$]} \\
\midrule
\multirow{8}{*}{USIS16K} & WaterMask R-CNN~\cite{lian2023watermask}$^{\ddagger}$ & 76.98 & 89.84 & 83.76 & 69.26 & 74.10 & 84.17 &  \\
 & Clarity-Former~\cite{chen2026clarityformer}$^{\S}$ & 82.59 & 92.03 & 88.70 & 62.42 & 79.98 & 91.00 &  \\
 & DiveSeg~\cite{chen2026diveseg}$^{\ddagger}$ & 87.95 & 96.73 & 94.20 & 71.10 & 87.41 & 94.20 &  \\
 & Baseline$^{a}$ & 85.72 & 95.31 & 92.81 & 71.27 & 83.35 & 92.73 &  \\
 & \method{} & 87.94 & 96.01 & 93.77 & 74.07 & 85.16 & 93.84 & $-0.01$ {\scriptsize[$-0.81$, $+0.58$]} \\
 & \quad + CPC & \underline{90.03} & \textbf{98.16} & \textbf{95.89} & \underline{76.08} & \underline{88.81} & \underline{94.31} & $+2.08$\sig{} {\scriptsize[$+1.36$, $+2.51$]} \\
 & \methodhr{} & 88.18 & 95.66 & 93.61 & 74.29 & 85.69 & 93.94 & $+0.23$ {\scriptsize[$-0.57$, $+0.93$]} \\
 & \quad + CPC & \textbf{90.41} & \underline{97.98} & \underline{95.82} & \textbf{78.40} & \textbf{88.92} & \textbf{94.45} & $+2.45$\sig{} {\scriptsize[$+1.73$, $+2.90$]} \\
\bottomrule
\end{tabular*}
\par\vspace{3pt}\begin{minipage}{\tblwidth}\raggedright\scriptsize One evaluator on the same test images. $\Delta$: our model minus the strongest competitor of the protocol (DiveSeg), with a paired image-bootstrap 95\,\% interval (600 draws); \sig{}: the interval excludes zero. $^{\S}$The authors' released weights; $^{\ddagger}$our retraining with the authors' code (one seed); on UIIS the released DiveSeg leads the retraining by $+2.30$\sig{} {\scriptsize[$+0.57$, $+4.26$]}. $^{a}$Base detector, three-seed mean. \textbf{Bold} / \underline{underlined}: best / second best. Size strata: mask areas at the COCO thresholds; USIS16K: mask area as a fraction of the image area ($<5$, $5$--$30$, $>30\,\%$). USIS10K-A: one category, no CPC.\end{minipage}
\end{table*}

Table~\ref{tab:same} removes split and evaluation differences: every method is scored by one evaluator on the same test images. DiveSeg, a ViT-L model with 390\,M parameters, is the strongest competitor on all five protocols. \method{} with CPC leads it on every protocol, with intervals that exclude zero: by 2.33 AP on UIIS, 5.94 on UIIS10K, 5.09 on USIS10K-A (one category, so without CPC), 7.13 on USIS10K-M, and 2.08 on USIS16K. On UIIS, the smallest test set, the margin is narrow: the lower end of the interval is $+0.09$. \methodhr{} with CPC leads by 2.45 to 6.55. Without the category branch, \method{} leads by 5.09 to 6.69 on UIIS10K and both USIS10K protocols and by 2.14 [$-0.04$, $+4.41$] on UIIS, and it ties DiveSeg on USIS16K ($-0.01$ [$-0.81$, $+0.58$]). On that 158-class taxonomy, CPC adds 2.09 AP to the same masks and separates the two methods. With CPC, \method{} is ahead of every competing method in every size stratum of every protocol, narrowly for large objects on UIIS and USIS16K (by 0.05 and 0.11 AP).

Where published scores exist, our scores of the competing methods match them within 1.0 AP or exceed them, with four exceptions. The released weights give 26.24 (WaterMask R-CNN) and 35.71 (DiveSeg) on UIIS, 65.62 (DiveSeg) on USIS10K-A, and 82.59 (Clarity-Former) on USIS16K, against published 27.2, 35.6, 64.1, and 82.5. Our retrained WaterMask R-CNN reaches 61.00, 38.98, and 76.98 on USIS10K-A, USIS10K-M, and USIS16K against published 59.0, 38.7, and 72.7, and our retrained Clarity-Former 28.84 on UIIS against 28.9. The exceptions fall short by 1.3 to 3.9 AP: the released DiveSeg on USIS10K-M (46.27 against 48.4), our WaterMask R-CNN on UIIS10K (36.07 against 37.4), our seven-class Clarity-Former on USIS10K-M (41.89 against 45.8, a score whose label space its paper does not state), and our DiveSeg on UIIS (33.41 against 35.6, and 2.30 [$+0.57$, $+4.26$] below the released weights). Our retrained DiveSeg models on UIIS10K and USIS16K may therefore understate the authors' own training by about 2.3 AP. On UIIS10K the lead of 5.94 AP is well beyond that; on USIS16K the lead of 2.08 is about as large, so it is the least certain of the five leads. DiveSeg's own TensorRT engine, which reads a fixed $480\times640$ input within its training scales, scores 88.84 on USIS16K, 0.89 above its result in Table~\ref{tab:same}; against it \method{} with CPC leads by 1.2 AP. Against the published DiveSeg scores, \method{} with CPC leads by 2.45 on UIIS and 5.00 on USIS10K-M.

\subsection{Efficiency and deployment}
\label{sec:exp-eff}

\begin{table*}[width=\textwidth,cols=9,pos=t]
\caption{Cost and end-to-end latency on the same test images.}\label{tab:eff}
\scriptsize
\setlength{\tabcolsep}{3pt}
\begin{tabular*}{\tblwidth}{@{\extracolsep{\fill}}llrrrrrrr@{}}
\toprule
 & & & \multicolumn{3}{c}{UIIS10K test images (0.31\,MP)} & \multicolumn{3}{c}{USIS16K test images (2.08\,MP)} \\
\cmidrule(lr){4-6}\cmidrule(l){7-9}
Method & Precision & Params (M)$\downarrow$ & ms (median / p90)$\downarrow$ & GFLOPs$\downarrow$ & AP$\uparrow$ & ms (median / p90)$\downarrow$ & GFLOPs$\downarrow$ & AP$\uparrow$ \\
\midrule
\multicolumn{9}{@{}l}{\textit{Ours: complete models on TensorRT (detector + W, P, L + PBR + Q + CPC)}} \\
\quad \method{} & FP32 & 120.3 & 66.3 / 77.7 & 327 & 49.91 & 52.4 / 65.5 & 235 & 90.02 \\
\quad \method{} & FP16 & 120.3 & 37.4 / 49.7 & 327 & 49.84 & 25.7 / 39.8 & 236 & 90.06 \\
\quad \quad + cheap PBR & FP16 & 120.3 & 28.1 / 36.0 & 264 & 49.82 & 24.9 / 46.0 & 226 & 89.93 \\
\quad \quad + cheap PBR + DINOv2-S & FP16 & 55.6 & 22.4 / 31.1 & 162 & 49.72 & 18.1 / 32.3 & 124 & 89.47 \\
\quad \methodhr{} & FP32 & 120.8 & 99.1 / 113.0 & 410 & 50.02 & 90.0 / 102.8 & 321 & 90.41 \\
\quad \methodhr{} & FP16 & 120.8 & 45.3 / 57.9 & 410 & 50.09 & 32.9 / 47.4 & 321 & 90.41 \\
\midrule
\multicolumn{9}{@{}l}{\textit{Competing methods: their TensorRT network with their own pre- and post-processing}} \\
\quad WaterMask R-CNN (R101) & FP16 & 67.5 & 53.3 / 65.1 & 420 & 36.07 & 93.9 / 175.3 & 1{,}052 & 76.98 \\
\quad Clarity-Former (Swin-S) & FP32 & 66.7 & 430.4 / 444.7 & 762 & 40.80 & 442.4 / 603.8 & 762 & 82.59 \\
\quad DiveSeg (ViT-L) & FP16 & 390.2 & 291.2 / 299.9 & 2{,}775 & 43.98 & 275.7 / 448.2 & 2{,}775 & 87.95 \\
\bottomrule
\end{tabular*}
\par\vspace{3pt}\begin{minipage}{\tblwidth}\raggedright\scriptsize One NVIDIA T4, batch 1, clocks not locked. ms: median / 90th percentile (p90) over 200 test images (images 11 to 210 of the test split, after 10 warm-up images), from upload of the decoded image to final outputs in host memory, without the shared JPEG decode (2.6--4.0 and 6.4--10.2\,ms). Precision: of the detector (refiner and crop encoder: FP16). GFLOPs: per image, mean over the timed images. AP: test AP of the protocol; for the competing methods, that of their PyTorch pipeline (Table~\ref{tab:same}); their timed engines score 34.00 (WaterMask R-CNN), 39.94 (Clarity-Former), and 44.36 (DiveSeg) on UIIS10K and 76.19, 79.23, and 88.84 on USIS16K. Params: ours with the 158-class probe; competing methods with the larger of their two models. FP32 rows: the models evaluated in Tables~\ref{tab:sota} and~\ref{tab:same}; the FP16 rows below them are faster operating points.\end{minipage}
\end{table*}

Table~\ref{tab:eff} and Fig.~\ref{fig:teaser} compare every method on one GPU model under one latency protocol, each competing method with the model trained for the protocol of the test images. End to end, the FP32 pipeline takes 66.3\,ms per image on the UIIS10K test images and 52.4\,ms on the USIS16K test images, which average 2.08 megapixels (MP). With the detector in FP16, which changes test AP by at most 0.14 against Table~\ref{tab:same} on all five protocols, the times fall to 37.4 and 25.7\,ms (40.3 and 33.6\,ms with JPEG decoding). Against each competing method's TensorRT pipeline, this is 1.4 times faster than WaterMask R-CNN, whose UIIS10K model reads 512-pixel inputs, 7.8 times faster than DiveSeg, and 11.5 times faster than Clarity-Former on UIIS10K, and 3.6, 10.7, and 17.2 times faster on USIS16K. Clarity-Former's own PyTorch release is faster than its FP32 engine (335.2 and 358.7\,ms from the image file). Including JPEG decoding for all methods and comparing with each competitor's faster pipeline, \method{} is 1.4 and 3.0 times faster than WaterMask R-CNN on UIIS10K and USIS16K, 7.3 and 8.4 times faster than DiveSeg, and 8.3 and 10.7 times faster than Clarity-Former. In every one of these comparisons \method{} is also the more accurate method, by 1.2 to 15.8 AP against the AP of the pipeline that was timed. A refiner setting that skips detections scoring below 0.10 and refines at most 32 patches per mask, and a DINOv2-S crop encoder, reach 22.4 and 18.1\,ms at a cost of 0.2 and 0.6 AP relative to the FP32 pipeline on these two protocols. \methodhr{} needs 1.2 to 1.7 times \method{}'s end-to-end time for its average edge of 0.25 AP. The complete model needs 235 to 327\,GFLOPs per image, against 420 to 1{,}052 for WaterMask R-CNN, 762 for Clarity-Former, and 2{,}775 for DiveSeg (Eq.~(\ref{eq:cost})); Supplementary Table~S1 adds the latency of each network engine alone. On the T4, the FP16 pipeline with CPC needs at most 1.6\,GiB of GPU memory on the UIIS10K test images and 2.5\,GiB on the USIS16K images, the CUDA context included, and its engines take 229\,MiB on disk.

\subsection{Ablation studies}
\label{sec:exp-ablation}

\subsubsection{Design of the DGC detector}
\label{sec:abl-dgc}

\begin{table*}[width=\textwidth,cols=8,pos=t]
\caption{Design of the DGC detector (mask AP, \%).}\label{tab:dgc}
\footnotesize
\setlength{\tabcolsep}{4pt}
\begin{tabular*}{\tblwidth}{@{\extracolsep{\fill}}lcccrrlr@{}}
\toprule
Config. & Attn. & Frame & Canvas & UIIS10K$\uparrow$ & USIS10K-A$\uparrow$ & $\Delta$ (pooled)$\uparrow$ & GFLOPs$\downarrow$ \\
\midrule
Baseline & \xmark & \xmark & \xmark & 46.04 & 68.04 & reference & 64.4 \\
attention only & \cmark & \xmark & \xmark & 45.93 & 68.26 & $+0.06$ {\scriptsize[$-0.25$, $+0.35$]} & 75.2 \\
frame only & \xmark & \cmark & \xmark & 46.47 & 68.64 & $+0.52$\sig{} {\scriptsize[$+0.19$, $+0.81$]} & 72.8 \\
canvas only & \xmark & \xmark & \cmark & 45.57 & 67.23 & $-0.64$\sig{} {\scriptsize[$-0.92$, $-0.36$]} & 76.1 \\
attention + canvas & \cmark & \xmark & \cmark & 46.13 & 67.80 & $-0.07$ {\scriptsize[$-0.40$, $+0.28$]} & 86.9 \\
frame + canvas & \xmark & \cmark & \cmark & 45.95 & 68.16 & $+0.01$ {\scriptsize[$-0.31$, $+0.31$]} & 84.5 \\
all three & \cmark & \cmark & \cmark & 46.40 & 68.53 & $+0.42$\sig{} {\scriptsize[$+0.05$, $+0.78$]} & 95.2 \\
\textbf{DGC} & \cmark & \cmark & \xmark & 46.82 & 68.89 & $+0.81$\sig{} {\scriptsize[$+0.45$, $+1.10$]} & 83.5 \\
\midrule
576-px input (HR) & \multicolumn{3}{c}{base detector, 576\,px} & 47.31 & 69.78 & $+1.51$\sig{} {\scriptsize[$+1.08$, $+1.88$]} & 166.9 \\
\bottomrule
\end{tabular*}
\par\vspace{3pt}\begin{minipage}{\tblwidth}\raggedright\scriptsize 20 epochs, three seeds per configuration, native detectors (without post-training components), official test. Attn.: one global attention window; Frame: supervision and mask decoding in the 576-px frame (backbone input 384\,px); Canvas: a dense 2$\times$ feature broadcast; \cmark: with the component, \xmark: without; HR: the base detector reading 576-px inputs (the detector of \methodhr{}). Per-protocol columns are three-seed means; $\Delta$ pools both protocols and seeds against the baseline, with the paired image-bootstrap 95\,\% interval (600 draws); \sig{}: the interval excludes zero. GFLOPs: the detector per image at its input.\end{minipage}
\end{table*}

Table~\ref{tab:dgc} varies three factors with the full training schedule: one global attention window, the 576-pixel supervision and decoding frame, and a dense canvas that broadcasts every backbone feature map onto a doubled grid. The frame carries most of the gain: alone it adds 0.52\sig{} [$+0.19$, $+0.81$] over the baseline, whereas global attention alone adds 0.06 at 384 pixels. Within the 576-pixel frame, global attention adds a further 0.30 [$-0.07$, $+0.60$], and the frame adds 0.76\sig{} [$+0.47$, $+1.00$] once global attention is present. Together they give the DGC detector, $+0.81$\sig{} over the baseline, which is 54\,\% of the gain of reading 576-pixel inputs ($+1.51$\sig{}) for 19\,\% of the FLOPs they add (19.1 against 102.5\,GFLOPs). Their interaction, $+0.24$ [$-0.14$, $+0.62$], has an interval that includes zero, so the design rests on the two effects themselves: a larger supervision and decoding frame, and global context computed on the compact grid. The dense canvas lowers pooled AP in all four configurations it enters, with intervals that exclude zero in three: it changes AP by $-0.64$\sig{} alone, by $-0.50$\sig{} [$-0.77$, $-0.21$] with the 576-pixel frame, by $-0.39$\sig{} in the full configuration, and by $-0.13$ with global attention alone.

At 12\,\% fewer FLOPs (83.5 against 95.1), the DGC detector also matches or exceeds the base detector run at 456 pixels on four of the five protocols: it is ahead on both USIS10K protocols ($+0.56$\sig{} and $+0.97$\sig{}), level within the intervals on UIIS10K and USIS16K, and behind only on UIIS, with an interval that includes zero ($-1.03$ [$-2.07$, $+0.01$]); pooled over the five protocols the difference is $+0.25$ [$-0.05$, $+0.57$]. With our post-training components applied to both, the DGC detector leads its dense-canvas variant on all five protocols (pooled $+0.38$ [$+0.09$, $+0.66$]) at 86\,\% of its detector engine time, and it comes within 0.25 AP of \methodhr{} on average (pooled $-0.25$ [$-0.75$, $+0.25$]) at 59\,\% of its detector engine time; both pooled intervals are normal approximations from the per-protocol intervals. Only on USIS10K-A is \methodhr{} ahead with an interval that excludes zero, by 0.87 AP. Where it leads, its advantage lies mainly in small and medium objects, which profit most from reading more pixels; on USIS10K-M, where \method{} is ahead overall, it also leads in every size stratum (Table~\ref{tab:same}).

\subsubsection{Post-training components}
\label{sec:abl-stack}

\begin{table*}[width=\textwidth,cols=12,pos=t]
\caption{The proposed post-training components of \method{} (mask AP, \%, official test).}\label{tab:stack}
\scriptsize
\setlength{\tabcolsep}{3pt}
\begin{tabular*}{\tblwidth}{@{\extracolsep{\fill}}lccccccrrrrrr@{}}
\toprule
Row & W & P & L & PBR & Q & CPC & UIIS$\uparrow$ & UIIS10K$\uparrow$ & USIS10K-A$\uparrow$ & USIS10K-M$\uparrow$ & USIS16K$\uparrow$ & Mean$\uparrow$ \\
\midrule
\multicolumn{13}{@{}l}{\textit{Cumulative}} \\
native (three-seed mean) & \xmark & \xmark & \xmark & \xmark & \xmark & \xmark & 36.23 & 46.82 & 68.89 & 50.92 & 86.82 & 57.94 \\
+ weight soup & \cmark & \xmark & \xmark & \xmark & \xmark & \xmark & 36.47 & 47.92 & 69.40 & 51.91 & 87.18 & 58.58 (+0.64\sig{}) \\
+ proposal selection & \cmark & \cmark & \xmark & \xmark & \xmark & \xmark & 36.60 & 47.99 & 69.45 & 52.15 & 87.21 & 58.68 (+0.10\sig{}) \\
+ layer averaging & \cmark & \cmark & \cmark & \xmark & \xmark & \xmark & 36.85 & 48.10 & 69.55 & 52.12 & 87.21 & 58.77 (+0.09) \\
+ boundary refiner & \cmark & \cmark & \cmark & \cmark & \xmark & \xmark & 37.23 & 49.41 & 69.87 & 52.39 & 87.46 & 59.27 (+0.51\sig{}) \\
+ quality = \textbf{Full} & \cmark & \cmark & \cmark & \cmark & \cmark & \xmark & 37.85 & 49.80 & 70.70 & 52.95 & 87.94 & 59.85 (+0.58\sig{}) \\
+ CPC & \cmark & \cmark & \cmark & \cmark & \cmark & \cmark & 38.05 & 49.92 & 70.70$^{b}$ & 53.40 & 90.03 & 60.42 (+0.57) \\
\midrule
\multicolumn{13}{@{}l}{\textit{Leave one out from Full ($\Delta$ AP)}} \\
$-$ weight soup & \xmark & \cmark & \cmark & \cmark & \cmark & \xmark & $-0.95$\sig{} & $-1.25$\sig{} & $-0.75$\sig{} & $-1.10$\sig{} & $-0.37$\sig{} & $-0.88$\sig{} \\
$-$ proposal selection & \cmark & \xmark & \cmark & \cmark & \cmark & \xmark & $-0.43$\sig{} & $-0.13$ & $-0.03$ & $-0.07$ & $+0.01$ & $-0.13$\sig{} \\
$-$ layer averaging & \cmark & \cmark & \xmark & \cmark & \cmark & \xmark & $-0.40$ & $-0.05$ & $-0.01$ & $+0.13$ & $0.00$ & $-0.07$ \\
$-$ boundary refiner & \cmark & \cmark & \cmark & \xmark & \cmark & \xmark & $-0.54$\sig{} & $-1.43$\sig{} & $-0.44$\sig{} & $-0.40$\sig{} & $-0.34$\sig{} & $-0.63$\sig{} \\
$-$ quality & \cmark & \cmark & \cmark & \cmark & \xmark & \xmark & $-0.63$ & $-0.38$\sig{} & $-0.83$\sig{} & $-0.57$\sig{} & $-0.48$\sig{} & $-0.58$\sig{} \\
Q on the unrefined mask & \cmark & \cmark & \cmark & \cmark & $\circ$ & \xmark & $-0.17$ & $-0.12$\sig{} & $-0.11$\sig{} & $-0.11$ & $-0.08$\sig{} & $-0.12$\sig{} \\
maskness instead of Q & \cmark & \cmark & \cmark & \cmark & $\diamond$ & \xmark & $+0.03$ & $-0.11$\sig{} & $-0.13$\sig{} & $-0.23$ & $-0.16$\sig{} & $-0.12$\sig{} \\
\bottomrule
\end{tabular*}
\par\vspace{3pt}\begin{minipage}{\tblwidth}\raggedright\scriptsize \method{}'s detector. W: weight soup of the three seed models; P: proposal selection; L: layer averaging; PBR: boundary refiner; Q: quality rescoring; CPC: category refinement; \cmark: applied, \xmark: not applied. Mean: over the five protocols; in the cumulative rows, the change from the previous row follows in parentheses, with \sig{} on the pooled change only; leave-one-out rows: difference from Full; \sig{}: the paired image-bootstrap 95\,\% interval excludes zero. $\circ$: Q on the unrefined mask. $\diamond$: the mean posterior inside the mask (maskness) in place of Q. $^{b}$USIS10K-A: one category, no CPC.\end{minipage}
\end{table*}

Table~\ref{tab:stack} adds the proposed post-training components one at a time to \method{}'s own detector and then removes one component at a time from Full. From the mean of the three native seed models to Full, AP rises by 1.12 (USIS16K) to 2.98 (UIIS10K), and by $+1.91$ [$+1.64$, $+2.10$] pooled over the five protocols. The weight soup, the refiner, and Q carry 1.73 of these 1.91 AP. The weight soup adds 0.64 AP pooled, the largest single step, with intervals that exclude zero on every protocol except UIIS: weight averaging, established for fine-tuned classifiers, also lifts this detection transformer, because its seed models keep their query slots aligned (next paragraph). The boundary refiner adds 0.51 pooled, from 0.25 on USIS16K to 1.32 on UIIS10K, with intervals that exclude zero on all five protocols. The quality score adds a further 0.58, with intervals that exclude zero on four protocols. Proposal selection, applied without the training that DDQ pairs it with, adds 0.10 AP pooled at no training cost, with an interval that excludes zero; layer averaging adds 0.09, with an interval that includes zero. Removing one component at a time from Full ranks the components in the same order, except that the refiner and Q swap places (0.63 and 0.58 AP pooled). The same leave-one-out matrix on the 576-pixel, dense-canvas, and baseline detectors (Supplementary Table~S1) agrees: removing the weight soup, Q, or PBR costs 0.42 to 0.95 AP pooled for each of them, while the intervals of P and L include zero.

Removing the soup costs the most, 0.88 AP pooled, with intervals that exclude zero on all five protocols, and every seed model on its own falls below the soup (by 0.75 to 1.03 AP pooled). Each seed model moves its weights from the COCO checkpoint by 1 to 2\,\% of their norm in the backbone and by at most 6\,\% elsewhere, and each of the detector's 100 queries is closest to the same query in the other two seed models (cosine 1.00, against 0.01 for any other query), so the soup averages corresponding parameters. On validation, with the 576-pixel and dense-canvas detectors, the soup keeps 40 to 64\,\% of the gain of a three-model output ensemble at a third of its cost, and a seed model whose backbone is replaced by the soup's backbone beats the original seed model in 14 of 15 cases: most of the soup's gain lies in the shared representation.

Without the refiner, Full loses 0.34 to 1.43 AP, more than the refiner adds before Q enters, because Q is computed on the refined masks and ranks them better; computing $q$ on the unrefined mask instead costs 0.12 AP pooled. Dropping the second term of Q's denominator also costs 0.12 AP pooled, with intervals that exclude zero on three protocols: the mean posterior inside the mask, which SOLO and Mask2Former use, ranks the refined masks worse than Q because it ignores the posterior mass outside the mask. Among the true positives, Q on the refined masks follows the true mask IoU more closely (Spearman correlation 0.72 to 0.87 across the protocols) than Q on the unrefined masks (0.44 to 0.86) and far more closely than the category score (0.35 to 0.61).

\subsubsection{Boundary refiner}
\label{sec:abl-pbr}

\begin{table*}[width=\textwidth,cols=10,pos=t]
\caption{The boundary refiner on the official test split. Left: input ablation ($\Delta$AP from the full refiner). Right: gains when the refiner is fitted to other detectors.}\label{tab:pbr}
\scriptsize
\setlength{\tabcolsep}{2pt}
\begin{minipage}[t]{0.588\textwidth}\vspace{0pt}\centering
\begin{tabular*}{\linewidth}{@{\extracolsep{\fill}}lccccrrrrr@{}}
\toprule
Variant & RGB & Post. & Band & Resid. & UIIS$\uparrow$ & UIIS10K$\uparrow$ & USIS10K-A$\uparrow$ & USIS10K-M$\uparrow$ & USIS16K$\uparrow$ \\
\midrule
full & \cmark & \cmark & \cmark & \cmark & $+0.38$\sig{} & $+1.26$\sig{} & $+0.34$\sig{} & $+0.23$\sig{} & $+0.23$\sig{} \\
\midrule
$-$RGB & \xmark & \cmark & \cmark & \cmark & $-0.25$\sig{} & $-0.33$\sig{} & $-0.10$ & $+0.06$ & $-0.06$ \\
$-$band & \cmark & \cmark & \xmark & \cmark & $-0.03$ & $-0.01$ & $-0.05$ & $+0.03$ & $+0.04$ \\
$-$post. & \cmark & \xmark & \cmark & \cmark & $-0.53$\sig{} & $-0.64$\sig{} & $-0.09$ & $-0.14$ & $-0.11$\sig{} \\
binary & \cmark & $\circ$ & \cmark & \cmark & $-0.97$\sig{} & $-0.62$\sig{} & $-0.22$\sig{} & $-0.43$\sig{} & $-0.09$\sig{} \\
BPR-like & \cmark & $\circ$ & \cmark & \xmark & $+0.01$ & $-0.04$ & $+0.07$ & $-0.05$ & $-0.04$ \\
\bottomrule
\end{tabular*}
\end{minipage}\hfill\begin{minipage}[t]{0.402\textwidth}\vspace{0pt}\centering
\begin{tabular*}{\linewidth}{@{\extracolsep{\fill}}llrl@{}}
\toprule
Detector & Protocol & AP$\uparrow$ & + PBR$\uparrow$ \\
\midrule
WaterMask R-CNN$^{\ddagger}$ & UIIS10K & 36.07 & $+1.79$\sig{} {\scriptsize[$+1.60$, $+2.00$]} \\
DiveSeg$^{\ddagger}$ & UIIS10K & 43.98 & $+0.67$\sig{} {\scriptsize[$+0.54$, $+0.83$]} \\
DiveSeg$^{\S}$ & UIIS & 35.71 & $+0.27$\sig{} {\scriptsize[$+0.003$, $+0.48$]} \\
Clarity-Former$^{\S}$ & USIS16K & 82.59 & $+0.16$\sig{} {\scriptsize[$+0.07$, $+0.25$]} \\
DiveSeg$^{\S}$ & USIS10K-M & 46.27 & $-0.05$ {\scriptsize[$-0.20$, $+0.09$]} \\
WaterMask R-CNN$^{\ddagger}$ & USIS10K-M & 38.98 & $-0.05$ {\scriptsize[$-0.22$, $+0.08$]} \\
\bottomrule
\end{tabular*}
\end{minipage}
\par\vspace{3pt}\begin{minipage}{\textwidth}\raggedright\scriptsize Left: first row, the full refiner's gain over no refiner; other rows, the change from it ($-$: one input removed; binary: the binary mask as input; BPR-like: binary input and a direct mask output); \cmark: input used, \xmark: input removed; Post.: the mask posterior, input as a clipped logit ($\circ$: the binary mask instead); Band: the uncertainty band; Resid.: residual output (\xmark: direct mask prediction, as BPR~\cite{tang2021bpr}). Right: the unchanged configuration fitted to other detectors ($^{\ddagger}$our retraining, $^{\S}$the authors' released weights). Every refiner on the left is fitted with the production configuration on one set of the detector's training-split predictions; the production refiner and its refit differ by at most 0.06 AP. \sig{}: the paired bootstrap 95\,\% interval excludes zero.\end{minipage}
\end{table*}

Table~\ref{tab:pbr} (left) removes one input of the refiner (Eq.~(\ref{eq:pbr})) at a time; every variant is fitted with the same training configuration on the same patches and scored on the official test split. The full refiner gains 0.23 to 1.26 AP, with intervals that exclude zero on all five protocols: 1.26 on UIIS10K, 0.38 on UIIS, 0.34 on USIS10K-A, and 0.23 on USIS10K-M and USIS16K. Its gain rests on the detector's soft posterior: removing the posterior costs 0.09 to 0.64 AP, with intervals that exclude zero on UIIS, UIIS10K and USIS16K, and feeding the binary mask instead, while still predicting a residual, costs 0.09 to 0.97 AP, with intervals that exclude zero on all five protocols. The uncertainty band changes AP by at most 0.05. The RGB patch adds 0.33 AP on UIIS10K and 0.25 on UIIS, and changes AP by at most 0.10 on the three USIS protocols, with intervals that include zero. The same network with a binary input and a replacement output, the arrangement of BPR~\cite{tang2021bpr}, is within 0.07 AP of PBR with intervals that include zero. The gain therefore requires neither a high-resolution segmentation network, as in BPR, nor a prompted foundation model~\cite{lin2025samrefiner}: a 47.7\,k-parameter network on boundary patches suffices, and the residual form lets the refiner start as the identity. Fitted unchanged to the outputs of other detectors (Table~\ref{tab:pbr}, right), PBR raises WaterMask R-CNN by 1.79 AP and DiveSeg by 0.67 on UIIS10K, DiveSeg by 0.27 on UIIS, and Clarity-Former by 0.16 on USIS16K; on USIS10K-M it changes the AP of DiveSeg and of WaterMask R-CNN by $-0.05$, with intervals that include zero.

A refiner's gain depends more on the benchmark it was fitted on than on the detector. On validation, with the refiners of \methodhr{}'s detector, a refiner fitted on one benchmark and applied to another retains only 37 to 48\,\% of UIIS10K's own-refiner gain, while the two USIS benchmarks exchange refiners almost freely (retention 88 and 104\,\%); pooled over the six pairs, a foreign refiner changes AP by $-0.33$ [$-0.42$, $-0.24$] against the target's own. Within one benchmark, a refiner fitted on one detector matches another detector's own refiner on validation (119\,\% of its gain). The annotation conventions differ accordingly: UIIS10K's polygons are dense (41.9 vertices per 100 pixels of perimeter), those of USIS10K and USIS16K sparser (19.1 and 10.5). To isolate the convention on the same images, every UIIS10K polygon was simplified to USIS16K's density with the Douglas--Peucker algorithm~\cite{douglas1973peucker} (tolerance 0.99 pixels, fixed on the training polygons), and a second refiner was fitted on the same patches with the simplified targets. Scored against the original and the simplified test annotations, each refiner gains most under the convention it was fitted on: the dense refiner leads by 0.09 AP on the original annotations and the simplified one by 0.05 on the simplified annotations, a convention-match contrast of $+0.14$ [$+0.03$, $+0.22$] on test. Both refiners gain less against the simplified annotations (0.83 and 0.88 AP, against 1.26 and 1.17 on the original ones). The polygon density thus accounts for part of the transfer loss; the loss it reproduces on identical images, 0.05 to 0.09 AP on test, is smaller than the 0.33 AP that a refiner fitted on another benchmark loses on validation, so other differences between the benchmarks contribute as well. PBR is detector-agnostic and convention-specific (Supplementary Section~S1 relates this to work on annotation style).

\subsubsection{Category branch}
\label{sec:abl-cpc}

\begin{table*}[width=\textwidth,cols=9,pos=t]
\caption{Category update rules compared on the same crops and probe (mask AP, \%).}\label{tab:cpc}
\footnotesize
\setlength{\tabcolsep}{4pt}
\begin{tabular*}{\tblwidth}{@{\extracolsep{\fill}}lcccrrrrr@{}}
\toprule
Update & Keeps & Fuses & Gates & UIIS$\uparrow$ & UIIS10K$\uparrow$ & USIS10K-M$\uparrow$ & USIS16K$\uparrow$ & Mean$\uparrow$ \\
\midrule
Full (no update) & -- & -- & -- & 37.85 & 49.80 & 52.95 & 87.94 & 57.13 \\
product $s\cdot p$ (DCR) & \cmark & \xmark & \xmark & 35.39 & 46.66 & 50.24 & 89.54 & 55.46 \\
hard relabel & \xmark & \xmark & \xmark & 35.77 & 46.17 & 50.38 & 89.25 & 55.39 \\
soft fusion & \cmark & \cmark & \xmark & 37.99 & 49.42 & 51.99 & 90.03 & 57.36 \\
disagreement gate & \cmark & \xmark & \cmark & 38.05 & 49.92 & 53.40 & 89.31 & 57.67 \\
\textbf{CPC} (rule) & \cmark & $\bullet$ & $\bullet$ & 38.05 & 49.92 & 53.40 & 90.03 & 57.85 \\
\bottomrule
\end{tabular*}
\par\vspace{3pt}\begin{minipage}{\tblwidth}\raggedright\scriptsize \method{}, official test. All rules use the same three crops per image, frozen DINOv2-B features, and probe. $s$: the detector's category score; $p$: the probe's probability; every score keeps the quality factor $q$. Keeps: the detector's hypotheses are kept; Fuses: probe and head are combined; Gates: acts only on confident disagreements; \cmark: yes, \xmark: no, --: no update; $\bullet$: soft fusion on USIS16K, the gate elsewhere (chosen on validation). Mean: over the four protocols. Pooled, CPC leads the product rule by $+2.39$\sig{} {\scriptsize[$+1.82$, $+2.92$]} and hard relabelling by $+2.45$\sig{} {\scriptsize[$+1.69$, $+3.26$]} AP; \sig{}: the paired image-bootstrap 95\,\% interval excludes zero.\end{minipage}
\end{table*}

The activation rule compares two classifiers on validation objects matched to ground truth (Eq.~(\ref{eq:rule})). On the validation objects of \method{}'s detector, the crop probe out-classifies the detector head by 3.1 points of top-1 accuracy on USIS16K, so soft fusion is selected, and trails it by 1.1 points on UIIS10K and 2.0 on USIS10K-M, so the gate is selected; for the 576-pixel and dense-canvas detectors, the probe leads by 3.2 to 3.9 points on USIS16K and trails by 1.0 to 1.4 on the coarse protocols. The AP effects of soft fusion on six validation settings (protocol--detector pairs of the latter two detectors) were known when the rule was written, and the rule was fixed in advance with the prediction that the accuracy comparison would reproduce all six, which it does: soft fusion gained 1.91 and 2.15 validation AP on USIS16K and lost 0.48 to 2.28 on the coarse protocols. On test, all twelve increments of CPC over Full are positive for the three detectors: large on USIS16K (2.09 to 2.23, with intervals that exclude zero), moderate on USIS10K-M (0.39 to 0.72), and small on UIIS and UIIS10K (0.03 to 0.20): there the gate never lowers AP.

Eq.~(\ref{eq:gate-identity}) explains why the gate costs little where the probe is the weaker classifier. On the validation objects of the 576-pixel and dense-canvas detectors, with the flip-averaged probe used to develop the rule, the gate fires on 1.1 to 2.1\,\% of the objects of the coarse protocols; when it fires, the probe is right on 47.5 to 61.8\,\% and the head on 35.3 to 50.0\,\%, so top-1 accuracy moves by $-0.03$ to $+0.33$ points. On USIS16K it fires on 4.3 to 4.4\,\% of objects with 89 to 93\,\% precision and raises top-1 accuracy by 3.4 to 3.9 points. Table~\ref{tab:cpc} keeps the crops, features, and probe fixed and varies only the update rule. Multiplying each hypothesis by the probe probability, as in DCR~\cite{cheng2018dcr}, or relabelling each selected mask loses 2.1 to 3.6 AP on the coarse protocols, where the gated update (Eq.~(\ref{eq:gate-update})) gains 0.12 to 0.44; on USIS16K the soft update (Eq.~(\ref{eq:soft})) leads the product rule by 0.49. The value of the branch lies in its update rule: it keeps the detector's hypotheses, combines the probe's evidence with the head's, and, on the protocols where the probe is the weaker classifier, acts only on confident disagreements. Reading more crops adds little (Supplementary Table~S4): on validation, the update selected by the rule gains at most 0.07 AP more with all masks than with the top three, whereas soft fusion on the coarse protocols falls as more crops are read. On USIS16K, the one-view branch that reads the top three masks retains 95 to 96\,\% of the gain of a flip-averaged branch that reads all masks (15.5 per image on average), at 9.5\,ms of encoder time instead of 97\,ms.

\subsection{Qualitative comparison}
\label{sec:exp-qual}

\begin{figure*}[pos=t]
\centering
\includegraphics[width=\textwidth]{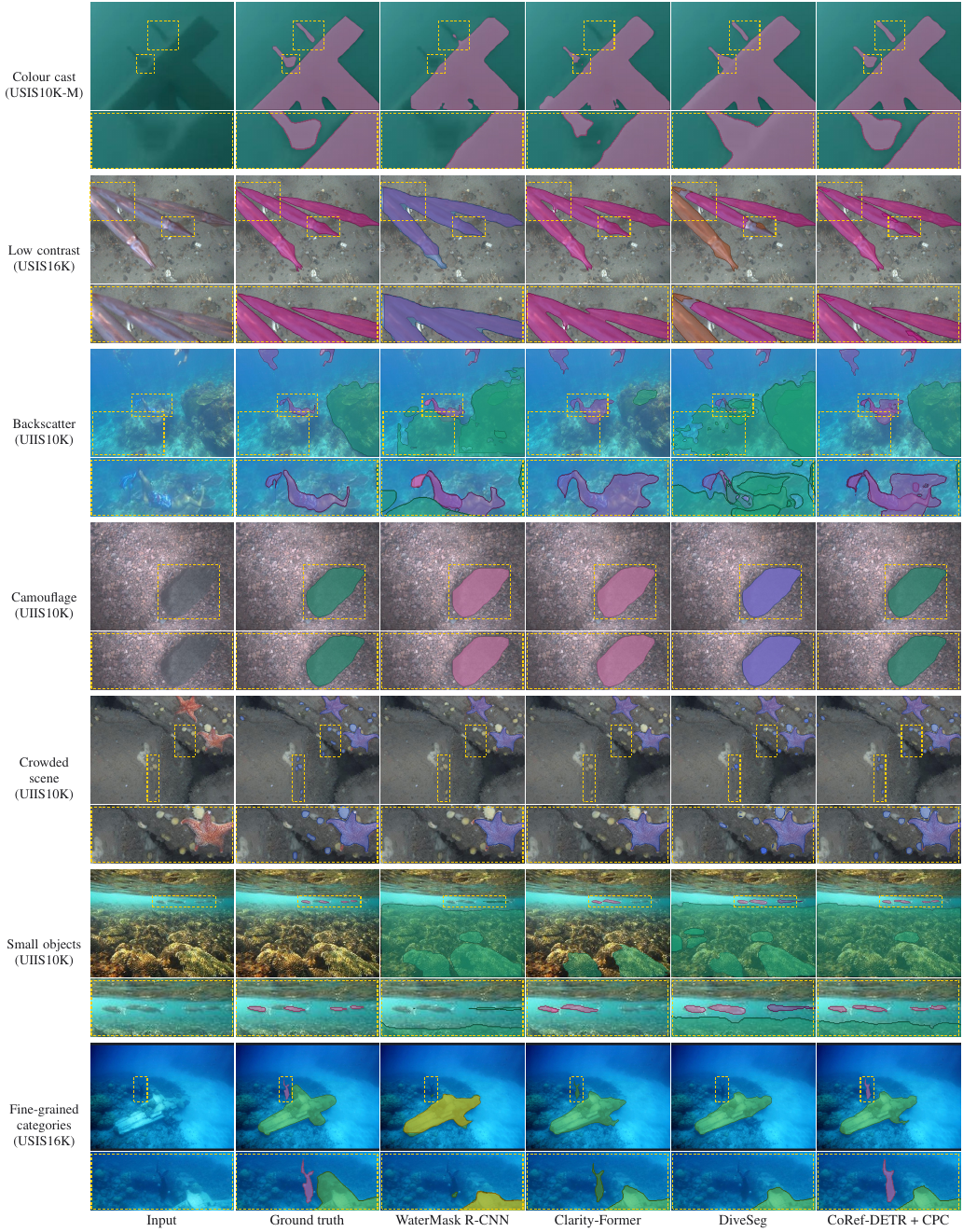}
\caption{Qualitative comparison on seven difficult scenarios from the official test partitions. Competing methods run the authors' released weights where they exist and our retraining otherwise (Table~\ref{tab:same}); every method is drawn with the same rule (distinct masks with score $\geq 0.3$, coloured by category). Dashed boxes mark regions where the methods differ, and the strip under each scenario enlarges the area around its first box. Supplementary Fig.~S1 adds random draws.}
\label{fig:qualgrid}
\end{figure*}

Fig.~\ref{fig:qualgrid} compares the methods on seven scenes, each showing one difficulty of underwater imagery. In the colour-cast scene, every method segments the wreck, but WaterMask R-CNN misses both small pieces of wreckage beside it, Clarity-Former finds only one, and DiveSeg merges one into the wreck; \method{} separates both. In the low-contrast scene, \method{} separates the three elongated animals with their annotated category; WaterMask R-CNN and Clarity-Former merge them into one mask, and DiveSeg labels one of them with another category. Under backscatter, \method{} returns the three divers and the coral on the right, while WaterMask R-CNN misses the distant diver on the left and spreads the coral over the sea floor, Clarity-Former segments only a fragment of the coral, and DiveSeg recovers only part of the nearest diver. In the camouflage scene, every method finds the flat object on the gravel, which the annotation labels as coral; WaterMask R-CNN and Clarity-Former call it a fish, DiveSeg returns overlapping fish and mollusc masks, and \method{} keeps the annotated category. In the crowded scene, every method segments the two sea stars, but WaterMask R-CNN and Clarity-Former miss all of the small shells around them, DiveSeg finds part of them, and \method{} finds nearly all. Around small objects, WaterMask R-CNN misses the row of small fish below the surface, which the other methods find, and every method also segments coral that the annotation leaves out. On the fine-grained USIS16K scene, \method{} is the only method that returns both the aircraft wreck and the diver with their annotated categories: WaterMask R-CNN labels the wreck with another category, Clarity-Former labels the diver with another category, and DiveSeg misses the diver. Supplementary Fig.~S2 shows what each stage of \method{} changes on test images.

\subsection{Failure analysis}
\label{sec:exp-fail}

\begin{table*}[width=\textwidth,cols=11,pos=t]
\caption{Where the remaining error lies: TIDE decomposition~\cite{bolya2020tide} of mask AP$_{50}$ (official test).}\label{tab:tide}
\footnotesize
\setlength{\tabcolsep}{4pt}
\begin{tabular*}{\tblwidth}{@{\extracolsep{\fill}}llrrrrrrrrr@{}}
\toprule
Protocol & Method & AP$_{50}{\uparrow}$ & Cls$\downarrow$ & Loc$\downarrow$ & Both$\downarrow$ & Dupe$\downarrow$ & Bkg$\downarrow$ & Miss$\downarrow$ & FP$\downarrow$ & FN$\downarrow$ \\
\midrule
\multirow{3}{*}{UIIS} & \method{} + CPC & 52.90 & 8.94 & 2.86 & 0.95 & 0.47 & 5.90 & 3.84 & 29.84 & 9.75 \\
 & \method{} & 52.66 & 9.02 & 2.84 & 0.96 & 0.45 & 5.91 & 3.83 & 30.08 & 9.72 \\
 & DiveSeg$^{\ddagger}$ & 49.99 & 4.59 & 4.14 & 1.41 & 0.34 & 8.07 & 4.88 & 31.62 & 9.87 \\
\midrule
\multirow{3}{*}{UIIS10K} & \method{} + CPC & 59.81 & 8.73 & 2.16 & 0.89 & 0.36 & 5.03 & 2.74 & 28.41 & 6.28 \\
 & \method{} & 59.64 & 8.76 & 2.14 & 0.92 & 0.35 & 5.04 & 2.71 & 28.57 & 6.24 \\
 & DiveSeg$^{\ddagger}$ & 56.58 & 8.38 & 3.32 & 0.78 & 0.60 & 4.43 & 3.10 & 28.67 & 8.24 \\
\midrule
\multirow{2}{*}{USIS10K-A} & \method{} & 88.87 & 0.00 & 1.57 & 0.00 & 0.26 & 5.75 & 0.34 & 9.15 & 1.37 \\
 & Clarity-Former$^{\ddagger}$ & 82.83 & 0.00 & 4.62 & 0.00 & 0.35 & 3.18 & 4.50 & 6.28 & 9.56 \\
\midrule
\multirow{3}{*}{USIS10K-M} & \method{} + CPC & 66.82 & 6.44 & 1.26 & 0.26 & 0.19 & 5.87 & 1.04 & 26.96 & 3.24 \\
 & \method{} & 66.34 & 6.75 & 1.27 & 0.27 & 0.21 & 5.86 & 0.96 & 27.44 & 3.16 \\
 & Clarity-Former$^{\ddagger}$ & 56.93 & 10.43 & 3.03 & 0.34 & 0.05 & 1.89 & 3.92 & 21.99 & 10.80 \\
\midrule
\multirow{3}{*}{USIS16K} & \method{} + CPC & 98.16 & 0.37 & 0.23 & 0.01 & 0.33 & 0.16 & 0.06 & 1.14 & 0.66 \\
 & \method{} & 96.01 & 2.58 & 0.17 & 0.02 & 0.04 & 0.17 & 0.06 & 3.12 & 0.81 \\
 & DiveSeg$^{\ddagger}$ & 96.73 & 1.03 & 0.43 & 0.02 & 0.17 & 0.36 & 0.36 & 2.24 & 0.91 \\
\bottomrule
\end{tabular*}
\par\vspace{3pt}\begin{minipage}{\tblwidth}\raggedright\scriptsize Each error column: the AP$_{50}$ gained by fixing that error type alone; FP and FN: by removing all false positives or recovering all false negatives. Cls: wrong category; Loc: right category, IoU 0.1 to 0.5; Both: both; Dupe: a second detection of a matched instance; Bkg: IoU below 0.1 with every instance; Miss: no detection covers the instance. $^{\ddagger}$Last row of each protocol: the strongest retrained competitor (as in Table~\ref{tab:same}); on UIIS and both USIS10K protocols it is 2.3 to 4.4 AP weaker than the released DiveSeg.\end{minipage}
\end{table*}

Table~\ref{tab:tide} decomposes the remaining error with TIDE~\cite{bolya2020tide}, which reports the AP$_{50}$ that each error type costs by correcting that type alone. The comparison uses the strongest retrained competitor of each protocol (DiveSeg on UIIS, UIIS10K, and USIS16K; Clarity-Former on both USIS10K protocols), which on UIIS and both USIS10K protocols is 2.3 to 4.4 AP weaker than the released DiveSeg of Table~\ref{tab:same}. Against it, \method{} makes smaller localisation and missed-instance errors on all five protocols, by 0.20 to 3.05 and 0.30 to 4.15 points, and its false negatives cost 0.12 to 8.19 points less. Two error types remain. Wrong categories are the largest error type on all four category-aware protocols (0.37 to 8.94 points), and on UIIS they cost more than DiveSeg's (8.94 against 4.59). Background detections cost 5.0 to 5.9 points on every protocol except USIS16K and exceed Clarity-Former's on both USIS10K protocols (5.75 against 3.18 and 5.87 against 1.89): \method{} recovers instances that the competing methods miss and also returns more detections that match no annotation. CPC targets the category error. On USIS16K it removes 2.21 of the 2.58 points that wrong categories cost \method{} (86\,\%), and on the coarse protocols it changes this error by 0.04 to 0.30 points.

The category errors concentrate on classes without a definite extent. Counted over all scored detections, 64.5\,\% of the classification errors on UIIS are confusions among reefs, aquatic plants, and sea-floor, and 90.5\,\% involve at least one of the three; on USIS10K-M, with the same taxonomy, 71.3\,\% involve one of them. On UIIS10K, 63.5\,\% involve corals, plants, or molluscs (a class whose annotations include sea urchins). These classes name stretches of reef, vegetation, and seabed whose extent is ambiguous, so a mask can overlap an annotated region and still carry a category that the annotation assigns to a neighbouring one.

\begin{figure*}[pos=t]
\centering
\includegraphics[width=\textwidth]{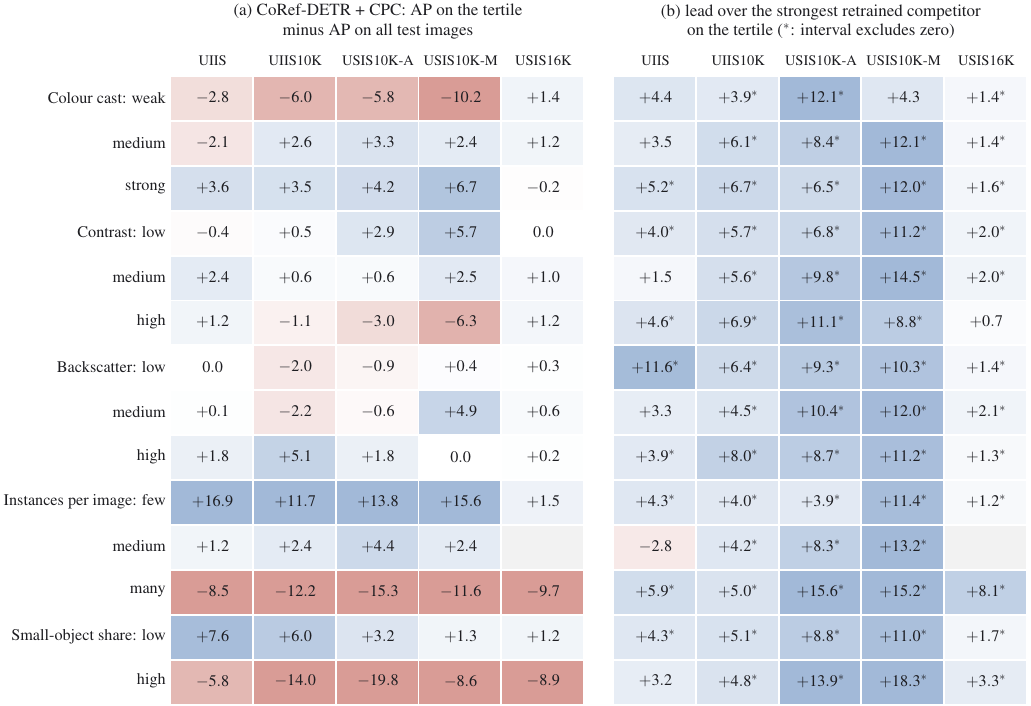}
\caption{Accuracy by image condition (official test). Each protocol's test images are split into tertiles of five measures (Section~\ref{sec:exp-fail}). (a) \method{} + CPC: mask AP on the tertile minus AP on all test images. (b) Its lead over the strongest retrained competitor ($^{\ddagger}$ in Table~\ref{tab:same}; on UIIS and both USIS10K protocols weaker than the released DiveSeg); $^{\ast}$: the paired bootstrap 95\,\% interval on the tertile excludes zero. The small-instance share has two strata (none, at least one). Blue: above zero; red: below; grey: empty stratum.}
\label{fig:strata}
\end{figure*}

Fig.~\ref{fig:strata} splits each test set into tertiles of five image properties: the colour cast $K$ in CIELab~\cite{gasparini2004colorbalancing}, the RMS contrast of $L^{\ast}$, backscatter as the mean underwater dark channel~\cite{drews2013udcp}, the number of instances, and the share of small instances (COCO small; on USIS16K, the benchmark's area stratum). Ties go to the lower tertile, so the small-instance share, which is zero for more than two thirds of the images of every protocol, has two strata, as does USIS16K's instance count. \method{} leads the strongest retrained competitor in 68 of the 69 tertiles, with an interval above zero in 61; the exception is the middle crowding tertile of UIIS, the smallest test set ($-2.80$ [$-10.29$, $+5.76$]). The number of instances and the share of small instances track \method{}'s accuracy most closely. From the tertile with the fewest instances to the one with the most, its AP falls by 23.9 to 29.2 points on four protocols and by 11.2 on USIS16K, and images with small instances score 9.9 to 23.0 points below images without them. Its lead grows with crowding on four protocols, from $+3.9$ to $+15.6$ on USIS10K-A and from $+1.2$ to $+8.1$ on USIS16K, so the gap to the competing methods is largest where all methods are weakest. The three degradation measures show no consistent direction. On four protocols the most colour-cast tertile scores above the least cast one, because these measures vary with scene content: on UIIS10K, the least cast tertile averages 5.5 instances per image and the most cast tertile 2.8. The figure therefore attributes no effect to colour cast, contrast, or backscatter as such; across the conditions that the four benchmarks contain, crowding and object size account for the largest differences in accuracy.

\begin{figure*}[pos=t]
\centering
\includegraphics[width=\textwidth]{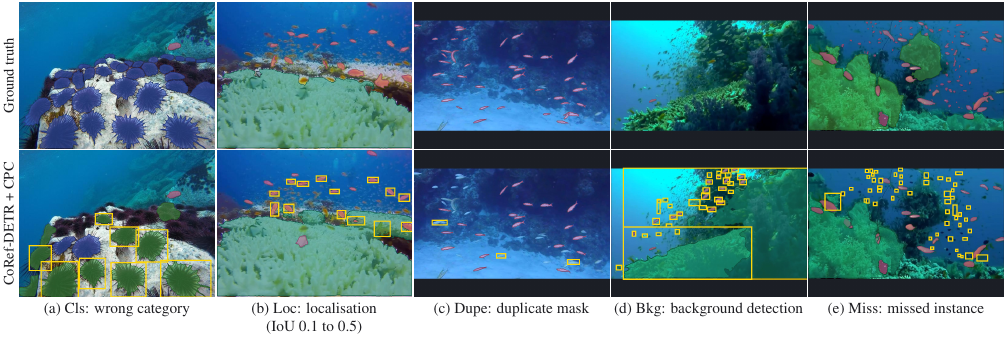}
\caption{Failure cases of \method{} + CPC: for five of the six TIDE error types (all but `both'), the test image with the most errors of that type (detections with score $\geq 0.3$; missed instances: all), chosen by a rule fixed in advance. Top: ground truth; bottom: predictions, with the errors of the column's type boxed in yellow; masks are coloured by category.}
\label{fig:failures}
\end{figure*}

Fig.~\ref{fig:failures} shows, for each error type, the test image on which \method{} makes the most errors of that type. All five come from UIIS10K, and four show schools of small fish: 14 masks of single fish that overlap an annotated group of fish with IoU 0.1 to 0.5, three duplicate masks, 35 detections that match no annotation, and 52 missed instances, 51 of them fish. In the classification case, \method{} labels eight sea urchins, which UIIS10K annotates as molluscs, as plants. The background case reflects the annotation as much as the model: its ground truth marks one fish of a large school, most of the 35 background detections fall on unannotated fish of that school, and the two largest cover unannotated coral. The remaining error of \method{} thus lies in crowds of small instances, in classes without a definite extent, in detections without a matching annotation, part of which mark objects the annotation omits, and in annotations that group several objects into one instance.
\section{Discussion}
\label{sec:discussion}

The results show that a compact detector can reach state-of-the-art accuracy on underwater benchmarks when its computation is spent on its three main sources of error: context, boundaries, and categories. For context, supervising and decoding masks in a larger frame brings most of the DGC detector's gain, and global attention on the compact token grid adds a smaller share; together they recover 54\,\% of the gain of 576-pixel inputs for 19\,\% of the FLOPs those inputs add. For boundaries, a refiner of only 47.7\,k parameters corrects the detector's own mask posterior, and the quality score is most effective when it is computed on the corrected masks. For categories, evidence from a frozen encoder pays off on fine-grained taxonomies, while the disagreement gate keeps it from hurting the coarse ones. Each stage adapts to the error patterns, taxonomies, and annotation conventions of the data. Across the conditions that the four benchmarks contain, crowding and object size account for the largest differences in accuracy (Section~\ref{sec:exp-fail}).

\method{} and \methodhr{} offer two operating points. \methodhr{} is 0.25 AP more accurate on average, and its detector engine takes 1.7 times as long as that of \method{}. The category branch adds 2.1 AP on the fine-grained USIS16K taxonomy through a frozen encoder of 86.6\,M parameters, and it can be omitted when fine-grained categories are not needed. For deployment, an FP16 detector nearly halves the end-to-end time while changing test AP by at most 0.14, and on the test images used for timing refining at most 32 patches per mask and encoding crops with DINOv2-S saves a further 30 to 40\,\% of the time for 0.1 to 0.6 AP.

Although \method{} achieves state-of-the-art accuracy on all five protocols, some limitations remain. First, the boundary refiner and the category probe are fitted for each benchmark, and the refiner works best on the annotation style it was fitted on (Section~\ref{sec:abl-pbr}). Second, all experiments train and test within the same benchmark, so robustness to new sites, cameras, and water conditions remains to be verified. Third, crowded scenes with many small instances remain the most difficult case for \method{}, as for the competing methods, although part of the measured error in these scenes comes from objects that the annotations omit (Section~\ref{sec:exp-fail}). Fourth, latency and memory were measured on a T4 GPU, where the pipeline needs at most 2.6\,GiB; its latency, memory, and power consumption on the embedded platforms of underwater vehicles have not yet been measured.
\section{Conclusion}
\label{sec:conclusion}

\method{} treats context, boundaries, and categories as three sources of error in a compact underwater instance segmenter and gives each its own stage. Computing global context on a compact token grid while supervising and decoding masks in a larger frame recovers 54\,\% of the gain of 576-pixel inputs for 19\,\% of the FLOPs they add, without new parameters. Most of this gain comes from the larger frame rather than from global attention. Correcting the detector's own mask posterior along boundaries, and ranking the corrected masks by their expected IoU, adds a 47.7\,k-parameter network and improves AP on all five protocols. The refiner's gain depends more on the benchmark it was fitted on than on the detector. A crop probe on frozen features adds 2.1 AP on the fine-grained USIS16K taxonomy, provided that a rule fixed on validation data keeps it from overriding the detector on coarse ones. Averaging the three models trained for the evaluation gives the largest post-training gain at no inference cost, because the seed models keep their query slots aligned. Across the four benchmarks, crowding and object size account for the largest differences in accuracy. Together, these stages set a new state of the art on all five protocols and outperform three reproduced recent underwater methods on the same test images (on USIS16K, with the category branch), while the TensorRT pipeline runs 1.4 to 11 times faster than these methods on a T4 GPU.

In future work, we plan to condition the boundary refiner on the annotation convention, so that one refiner can serve several benchmarks, and to make the category branch effective on coarse taxonomies as well, where the crop probe is currently a weaker classifier than the detector head. We also aim to improve the segmentation of small instances in crowded scenes, which remain the hardest cases for all methods, and to evaluate the model across datasets and on the embedded hardware of underwater vehicles.

\printcredits

\section*{Declaration of competing interest}
The authors declare that they have no known competing financial interests or personal relationships that could have appeared to influence the work reported in this paper.

\section*{Funding}
This research did not receive any specific grant from funding agencies in the public, commercial, or not-for-profit sectors.

\section*{Data availability}
The datasets are publicly available from their original publications~\cite{lian2023watermask,li2025uwsam,lian2024usissam,hong2025usis16k}. The code and trained weights will be publicly available at \url{https://github.com/mfazrinizar/CoRef-DETR} upon publication. The supplementary material, which details the implementation, will also be released upon publication.

\section*{Declaration of generative AI and AI-assisted technologies in the manuscript preparation process}
During the preparation of this work the authors used Claude and ChatGPT in order to assist with draft editing. After using these tools, the authors reviewed and edited the content as needed and take full responsibility for the content of the published article.

\bibliographystyle{elsarticle-num-names}
\bibliography{references}

\end{document}